\documentclass{article}

\usepackage{spconf}
\usepackage{amsmath}
\usepackage{amssymb}
\usepackage[pdftex]{graphicx}
\graphicspath{{./Figures/}}
\usepackage{multirow}
\usepackage{url}

\usepackage{dsfont}
\usepackage{url}
\usepackage{subfig}
\usepackage{xcolor}

\newcommand{\cbox}[1]{\raisebox{\depth}{\fcolorbox{black}{#1}{\null}}}
\newcommand{\argmax}{\arg\!\max}

\title{SPiKeS: Superpixel-Keypoints Structure for Robust Visual Tracking}
\name{Fran\c cois-Xavier Derue{$^{1}$}, Guillaume-Alexandre Bilodeau{$^{1}$}, Robert Bergevin{$^{2}$}}
\address{Polytechnique Montr\'eal{$^{1}$}, Universit\'e Laval {$^{2}$}}
\begin{document}
\maketitle
\begin{abstract}
In visual tracking, part-based trackers are attractive since they are robust against occlusion and deformation. However, a part represented by a rectangular patch does not account for the shape of the target, while a superpixel does thanks to its boundary evidence. Nevertheless, tracking superpixels is difficult due to their lack of discriminative power. Therefore, to enable superpixels to be tracked discriminatively as object parts, we propose to enhance them with keypoints. By combining properties of these two features, we build a novel element designated as a Superpixel-Keypoints structure (SPiKeS). Being discriminative, these new object parts can be located efficiently by a simple nearest neighbor matching process. Then, in a tracking process, each match votes for the target's center to give its location. In addition, the interesting properties of our new feature allows the development of an efficient model update for more robust tracking. According to experimental results, our SPiKeS-based tracker proves to be robust in many challenging scenarios by performing favorably against the state-of-the-art.
\end{abstract}


\section{Introduction}
\label{intro}

A robot needs to track its target to interact with it. Doubtful behaviors can be detected thanks to tracking in visual surveillance. Hands are tracked for gesture recognition. Those examples show only a small part of a wide range of tracking applications, thus encouraging many research efforts to focus on this topic. When no prior information about the object to track is available, tracking is referred to as model-free tracking. In a video, the goal is to locate a particular target given its location only in the first frame. This task is challenging because of numerous factors such as illumination variation modifying object color, occlusion hiding some parts, or new parts appearing if the viewpoint changes. While some of these issues are handled efficiently by different techniques, it is challenging for a single tracker to handle them all.

Trackers are generally split into two categories: discriminative and generative. Discriminative trackers consider tracking as a binary classification problem. Samples from foreground and background are selected to train a classifier that is able to separate the target from the rest of scene. Afterwards, this target detection yields a location estimation. This is the typical ``tracking-by-detection" framework followed by many discriminative trackers \cite{SVM_tracker,RET, MILT, OMSIT}, although Struck \cite{Struck} achieves the classification and location in one step. In most approaches, samples are selected randomly, limiting their number for computational efficiency. Instead of random samples, Henriques et al. \cite{CSK} proposed to select all the samples and exploited their redundancy to build a kernel classifier which tracks very quickly in the Fourier domain. Due to its simplicity and rapidity, many recent trackers build upon it \cite{CSK_color, KCF, RPT}.

 \begin{figure}[t]
\begin{center}
\includegraphics[width=1.0\linewidth]{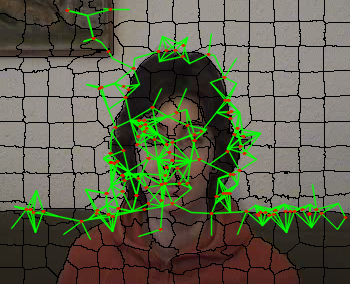}
\end{center}
\caption{Decomposition of a frame into SPiKeS. Superpixels (black) stuctured by keypoints (red dot) linked by vectors (green).}
\label{fig:fourFace}
\end{figure}
In generative trackers, only foreground information models the target appearance and the tracking task aims to find the most similar image region to this model. In \cite{L1}, the target is represented with a sparse model by using templates. The tracking location is the patch whose projection error into the template space is minimum. To account for appearance change and different kinds of motion, Kwon et al. \cite{VTD} build different observation and motion models, so that each pair can be used within a basic tracker. These multiple basic trackers are then integrated into a main tracker, which is more robust thanks to the interaction between its components. 

Although these methods can handle some appearance alterations, they are not robust against deformation and occlusion due to their holistic representation.
These issues are usually handled by the family of part-based trackers. As the model is decomposed into several parts, an occlusion only affects some of them, without preventing the other ones to track the target. Typically, usual approaches consider the parts as rectangular patches structured in a grid \cite{FragTrack,patch1,patch2,patch3}. However, non-rectangular targets are not well represented because background patches inside the bounding box inevitably affect the model and make it drift. To this end, Li et al. \cite{RPT} assigns reliability to patches so that noisy background patches do not affect the tracking. 

Another part-based approach consists of oversegmenting the target into superpixels. Thanks to their boundary evidence, they take better the shape into account. In \cite{SPT}, a map is built showing the probability of a superpixel to belong to the target and the target location is the area with maximum likelihood. This tracker shows good performance but it needs a model which has to be learned in the first frames. Therefore, it needs manual annotation or another tracker for the initialization step. Recent approaches such as \cite{DGT} and \cite{many2many} propose to integrate superpixels in a matching-tracking framework. An appearance model is built with superpixels and each of them attempts to find a match in the new frame in order to locate the target. One common problem is the low discriminative power of superpixel resulting in ambiguous matches. It then requires a complex strategy for matching.

 Better features for matching are the keypoints. Because of their saliency and invariance to transformations, a keypoint-based appearance model can be matched efficiently even in case of occlusion and deformation. Nonetheless, keypoint-based trackers \cite{kp_vote1,kp_vote2,kp_struck,wassim} often fail to represent uniform regions, where no keypoints can be found.
 
 Therefore, we hypothesize that superpixels and keypoints can complement each other. An object can always be segmented into superpixels but their lack of discriminative power makes them hard to match. Conversely, keypoints are more reliable to match but they poorly represent uniform-colored and non-textured regions. In our method, we propose to combine the assets of these two features in a single one: a Superpixel-Keypoints structure (SPiKeS). This is our first contribution. Figure \ref{fig:fourFace} illustrates a frame decomposed into SPiKeS. Notice that keypoints contributing to a SPiKeS can be inside the superpixel or nearby. A single keypoint can contribute to many SPiKeS. Incomplete SPiKeS are possible if there is no keypoint around. In that case, they are only described with the superpixel. 

Our second contribution is the design of a tracker that capitalizes on the SPiKeS. Experimental results show that our SPiKeS-based tracker performs well in numerous challenging situations and performs favorably against state-of-the-art trackers.

The paper is structured as follows. Section \ref{sec:relWork} presents works related to ours, i.e. trackers based on superpixels or keypoints. Our combination of these two features to build a SPiKeS is described in Section \ref{sec:description}. Then, section \ref{sec:tracker} shows how this new feature can be integrated into a tracking framework for robust target location estimation. Finally, the evaluation of section \ref{sec:experiment} compares the proposed tracker to the state-of-the-art.

\section{Related work} \label{sec:relWork}

The idea of combining keypoints with other features for tracking has been exploited in \cite{patch_sift}. The RGB and LBP histograms are extracted from patches to create an appearance model. SIFT keypoints \cite{SIFT} are then detected and their disposition is described by a circular histogram that represents a global geometric structure of the target. Our method is more flexible against deformation as each SPiKeS has its own keypoint structure, which allows local deformations. 

Instead of patches, Liu et al. \cite{Surf_Spx} proposed a tracker based on superpixels and SURF keypoints \cite{surf}. But unlike our proposition, their matching step only involves keypoints. The superpixels are only used for their boundary evidence, which is a useful clue when updating the model. Indeed, a new keypoint belonging to the same superpixel than a matching keypoint tends to be a part of the target since every point within a superpixel is likely to belong to the same object. We also consider this benefit in our method but in addition, because we are matching superpixels, even if there is no matching keypoint inside them, new keypoints can still be added. Therefore, our model update is more accurate.

Our localization process is inspired by \cite{kp_vote1,kp_vote2}. Their approach assigns to each matching keypoint a vote for the center of the target, allowing keypoints to locate the target independently from each other. Hierarchical clustering then converges to a consensus of votes such that outliers are removed. Finally, the selected votes estimate the position as a simple center of mass. Furthermore, Bouachir et al. \cite{wassim} proposed to weight the votes according to the reliability of keypoints. We do the same, but instead of voting with keypoints, we vote with SPiKeS.

\section{Superpixel-Keypoints Structure}\label{sec:description}

\begin{figure}[t]
\begin{center}
\includegraphics[width=0.7\linewidth]{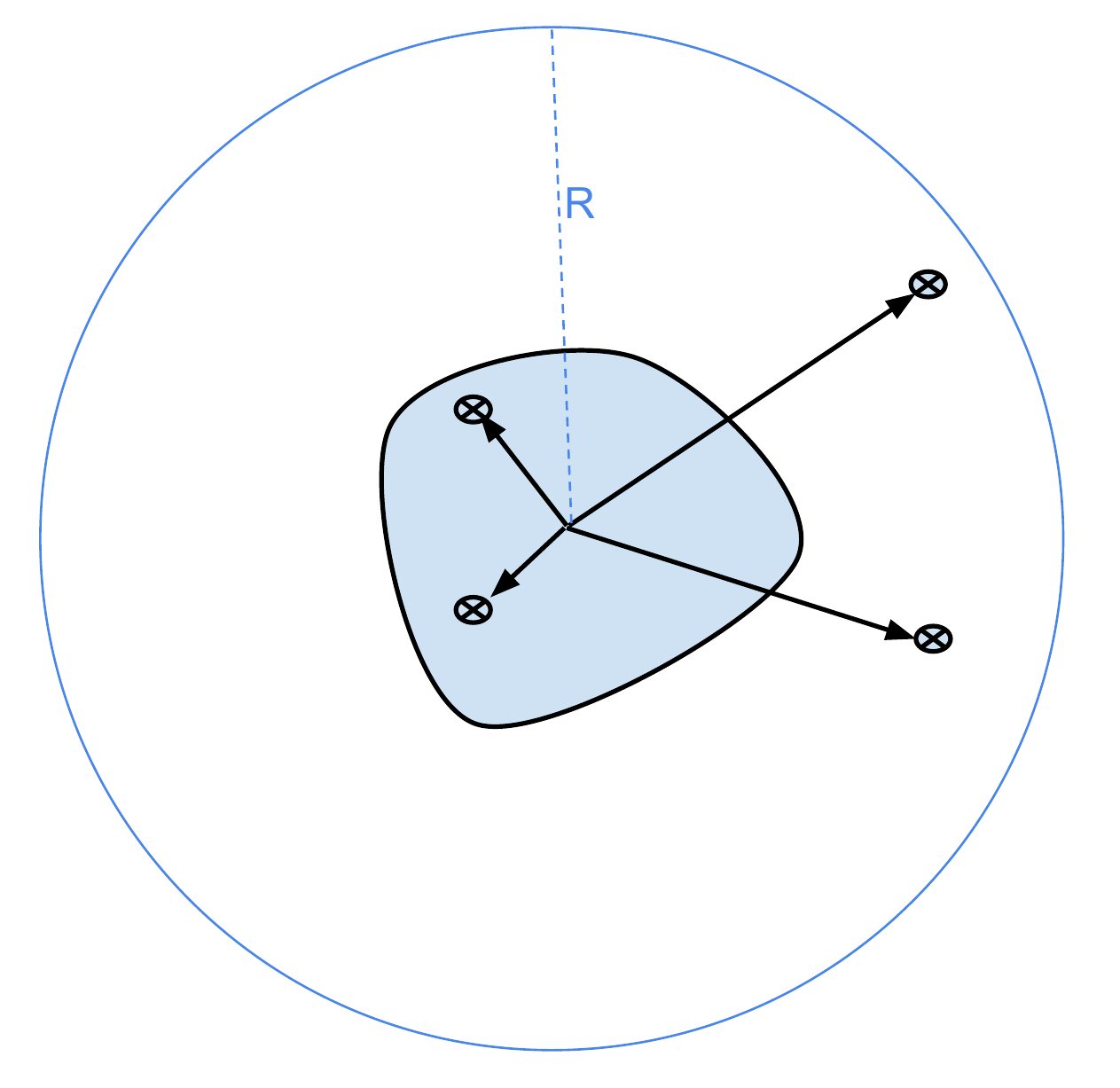}
\end{center}
\caption{SPiKeS representation. Keypoints are found inside or nearby the superpixel in a region of radius $R$ around its center. Keypoints relative positions are given by vectors.}
\label{fig:Spike}
\end{figure}

As shown in figure \ref{fig:Spike}, a Superpixel-Keypoints Structure ( SPiKeS ) consists in a superpixel and all the keypoints found in a region of radius $R$ around that superpixel's center. It implies that keypoints can be inside and outside the superpixel. Each keypoint is linked to the superpixel's center by a vector with a magnitude and an orientation. Therefore, a SPiKeS is a superpixel that acquired a spatial structure of keypoints, making it more discriminative. A SPiKeS without any keypoints is simply a superpixel.

\subsection{SPiKeS definition} \label{sec:def}

Let  $s$ be a superpixel and $\mathbf{k}$ the set of keypoints around $s$, we write the associated SPiKeS denoted by $S$ as
\begin{eqnarray}
S = \{(s,\mathbf{k})\mid \mathbf{k} = (k_1, ...,k_n,...,k_N),||\mathbf{x}_n^{k}{-}\mathbf{x}^{s}||{<}R\}
\end{eqnarray}
with $\mathbf{x}^{s}$ and $\mathbf{x}_n^{k}$ the centers of superpixel $s$ and keypoint $n$ respectively. $N$ is the total number of keypoints found in a description region of radius $R$ centered on $\mathbf{x}^{s}$.  Therefore, we define a descriptor for a SPiKeS  as being $f = \{h,\mathbf{d}^{k},\mathbf{e}\}$ with

\begin{itemize}
\renewcommand{\labelitemi}{$\bullet$} 
\item $h$: HSV histogram of $s$
\item $\mathbf{d}^{k}= \{d_1^k,...,d_n^k,...,d_N^k\}$ with $d_n^k$ the descriptor of $k_n$.
\item $\mathbf{e} = \{\vec{e}_1,...,\vec{e}_n,...,\vec{e}_N\}$ with $\vec{e}_n = \mathbf{x}_n^{k}-\mathbf{x}^{s}$ the vector from the superpixel's center to $k_n$ 
\end{itemize}

\subsection{SPiKeS comparison}

In order to compare two SPiKeS, we propose a measure of similarity based on their descriptors. Let $z(S_i,S_j)$ be the similarity score between SPiKeS $S_i$ and $S_j$. Because the color information of the superpixel and the keypoints structure are available, the score is a contribution of two terms: 
\begin{equation}\label{eq:z}
z(S_i,S_j)= \left\{\begin{array}{ll}
z_c + z_{k} & \mbox{if $d(h_i,h_j)<\theta_c$}\\
0 & \mbox{otherwise}
\end{array}\right.
\end{equation}
The first term $z_c$ represents the similarity between superpixel's color histograms
\begin{equation}\label{eq:zc}
z_c = \exp(-d(h_i,h_j)) 
\end{equation}
with $d(.,.)$ being the Bhattacharyya distance measure.

The second term $z_k$ represents the similarity between keypoints structure. The higher the number of matching keypoints between $\mathbf{k}_i$ and $\mathbf{k}_j$, the higher the score. Moreover, if both keypoints of a matching pair are positioned similarly with respect to their superpixel's center, the score should also increase. Thus we define
\begin{equation}
z_{k} = \sum_{k_m\in \mathbf{k}_i}\sum_{k_n\in \mathbf{k}_j}\gamma_{mn}p_{mn}
\end{equation}
with $p_{mn}= 1$ if $k_m$ and $k_n$ match, else $p_{mn}=0$. 
The factor $\gamma_{mn}$ weights the contribution of a keypoints match by comparing edges $e_m$ and $e_n$. We compute $\gamma_{mn}$ with the vector difference magnitude normalized by the diameter of the description region:
\begin{equation}
\gamma_{mn} = \exp\left(-\frac{||\vec{e_m'}-\vec{e_n'}||}{2R}\right)
\end{equation}

Note that to benefit from keypoints rotation invariance, $e_m'$ and $e_n'$ are the vectors $e_m$ and $e_n$ reoriented according to the principal orientations given by keypoints $k_m$ and $k_n$ respectively.  

Finally, the threshold $\theta_c$ ensures a minimum of color similarity to handle the case of wrong matching keypoints resulting in a high $z_k$.

\section{The SPiKeS Tracker}\label{sec:tracker}

\begin{figure*}[t]
\begin{center}
\includegraphics[width=1\linewidth]{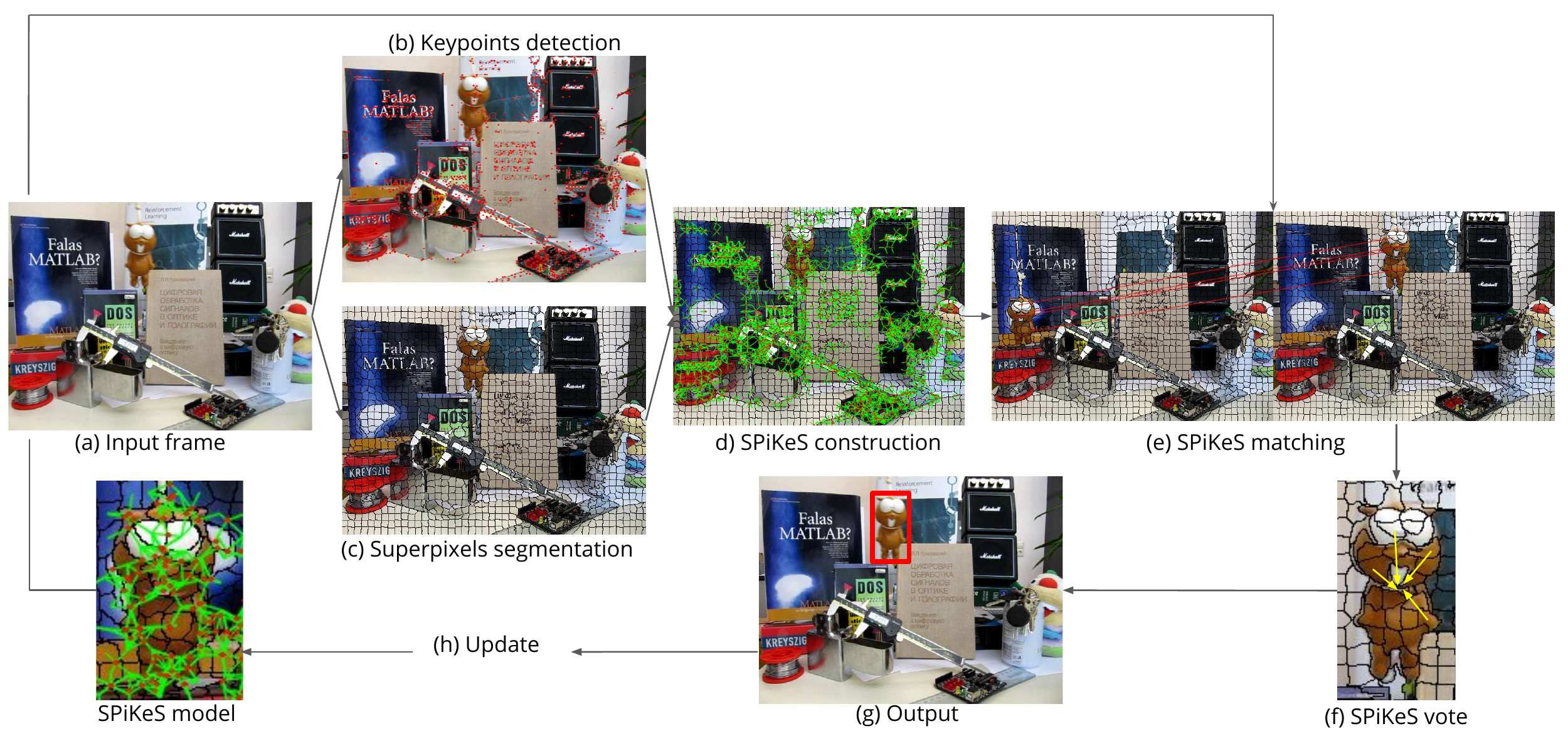}
\end{center}
\caption{Tracking steps of our SPiKeS-based tracker. Keypoints detection (b) and superpixels segmentation (c) are processed on an input frame (a). Each superpixel forms a SPiKeS with its surrounding keypoints (d). Our SPiKeS model is matched with the new SPiKeS and the matching ones vote for the target's center (f) . The model is updated from the estimated bounding box (g) if no occlusion occurs.}
\label{im:framework}
\end{figure*}

In model-free tracking, the information about the target location in the first frame is given by a bounding box. The SPiKeS are extracted from it to represent the appearance model. In the subsequent frames, after oversegmentation and keypoint detection, we build SPiKeS and locate those who match the model. Then, each matching SPiKeS votes for a position in the frame. The target's center is estimated from all votes. If no occlusion is detected, the model is updated. These tracking steps are illustrated in figure \ref{im:framework}.

\subsection{Model} \label{sec:model}
From the initial bounding box, we first detect keypoints, store them in a pool $\mathbf{K^f}$ and combine them with the $N_m$ extracted superpixels to form our model of SPiKeS: $\mathit{\mathbf{S}}^m = \{S_1^m,...,S_i^m,...,S^m_{N_m}\}$. Then, SPiKeS $S_i^m$ is assigned a vote vector $\mathbf{v}_i$ such that it can locate independently the target's center:
\begin{equation}
\mathbf{v}_i = \mathbf{x}_{0}^T-\mathbf{x}_{S_i^m}
\end{equation}
with $\mathbf{x}_{0}^T$ the target's center at time $t=0$, known as the center of the initial bounding box. We refer to $\mathbf{x}_{S_i^m}$ as the center of $S_i^m$ which is equivalent to its superpixel's center.

In addition, we extract keypoints in a surrounding region of the bounding box to keep a keypoint background model $\mathbf{K^b}$, which will help to detect occlusion similarly to \cite{muster}.


\subsection{Matching}

During tracking, we extract a pool $\mathbf{S}^q = \{ S_1^q,...,S_j^q,...S^q_{N_q} \}$ of $N_q$ SPiKeS from the entire incoming frame. Afterwards, we apply a greedy matching algorithm. The first step is looking for the nearest neighbour $S^q_{j^*}\in \mathbf{S}^q$ of every ${S}_i^m \in \mathbf{S}^m$:
\begin{equation}
(S_i^m,S_{j^*}^{q}) = \argmax\limits_j(z(S_i^m,S_j^q)) \quad \mbox{$i = 1,...,N_m$}
\end{equation}
However, $S_{j^*}^{q}$ could be the nearest neighbour of different $S_i^m$ meaning a many-to-one match. Since a one-to-one match is required, only the highest score is kept:
\begin{equation}
(S_{i^*}^m,S_{j^*}^{q}) = \argmax\limits_i(z(S_i^m,S_{j^*}^q)) 
\end{equation}

At this point, there are now $L\leq N_q$ one-to-one matches that we refer to as pairs of matches $M_l = (S_l^m,S_l^{q})$ with $l = 1,2,...,L$.

The next step consists in the rejection of wrong pairs of matches.
Firstly, a given SPiKeS of the model may not have a valid match in a given new frame, e.g. when a part is not visible. In this case, the nearest neighbour has a low matching score relative to a threshold and it can be discarded. We set a different value for the threshold according to the presence or not of matching keypoints. Indeed, if there are no keypoint matches, only the color provides the match between SPiKeS. As we already set a color threshold $\theta_c$, the minimum value of the matching score is $e^{-\theta_c}$ according to equation \ref{eq:z} and \ref{eq:zc}. On the other hand, if there are matching keypoints, the score will always be higher than this minimum value, thus the threshold is set higher.

Secondly, as we assume the target motion is smooth and continuous in time, a match $M_l$ is also considered inconsistent if the displacement between $S_l^m$ and $S_l^q$ is too large with respect to recent motion.

Formally, a matching pair $M_l$ is valid and not discarded if and only if  
\begin{equation} 
z(S_l^m, S_l^q)>\left\{\begin{array}{ll} 
e^{-\theta_c} & \mbox{if $N_{match}^{kp}=0$}\\ 
e^{-\theta_c} +\lambda_1& \mbox{otherwise} 
\end{array}\right. 
\end{equation} 
and 
\begin{equation} 
||\mathbf{x}_{S_l^{m}}-\mathbf{x}_{S_l^{q}}|| < ||\mathbf{x}_{t-1}-\mathbf{x}_{t-2}|| + \lambda_2
\end{equation} 
with $N_{match}^{kp}$ the total number of foreground keypoints matches, $\lambda_1$ a score threshold parameter and $\lambda_2$ a motion constraint parameter.


\subsection{Location Estimation}

Once the $L^*$ retained matching pairs have been determined, each $S_l^{q}$ votes for a position in the frame according to the vote vector $\mathbf{v}_{l}$ given by its respective $S_l^{m}$:
\begin{equation}
\mathbf{x}_l = \mathbf{x}_{S_l^{q}}+\mathbf{v}_{l}
\end{equation}

The estimated target location is computed by a weighted average of the votes: 
\begin{equation}
\mathbf{x}_t^T = \frac{\sum\limits_l^{L^*} \omega_{l} \phi_{l} \mathbf{x}_{l}}{\sum\limits_{l}^{L^*}\omega_{l}\phi_{l}}
\end{equation}
The factors $\omega_l$ and $\phi_l$, as introduced in \cite{wassim}, are the persistence and predictive factor of $S_l^{m}$. They give more importance to SPiKeS that often match and vote correctly for the target's center. More details are given in the next section.

\subsection{Update}
Section \ref{sec:model} introduced $\mathbf{K^f}$ and $\mathbf{K^b}$, our models of foreground and background keypoints. During the matching process, keypoints belonging to $\mathbf{K^b}$ are matched at the same time as the foreground ones. Once the new bounding box has been evaluated, if the number of keypoints inside it matching the background model exceeds a threshold $\theta_o$, an occlusion is detected. Therefore, no update takes place and the next frame is processed. Otherwise, if no occlusion occurs, the following update scheme is applied.

{\textbf{Step 1 :}} \textit{Descriptors and votes update.}
For each valid match $(S_{l}^{m},S_{l}^{q})$, the SPiKeS descriptor $f$ defined in subsection \ref{sec:def} is updated with:
\begin{equation}
f_{l}^m = (1-\alpha_f)f_{l}^m+\alpha_f f_{l}^q
\end{equation}
This simple formula adapts the model to gradual change of illumination by updating the color of the superpixels and the position and description of the keypoints.

For a non-rigid target, local parts tend to move with respect to the center. Vote vectors need to be modified according to the SPiKeS new position to take into account local deformations: 
\begin{equation}
\mathbf{v}_{l}^m = (1-\alpha_v)\mathbf{v}_{l}^m+\alpha_v (\mathbf{x}_t^T-\mathbf{x}_{S_{l}^q})
\end{equation}

As SPiKeS are the ``parts'' of our model, these terms are used interchangeably. A part that matches more often means that it is easier to identify and constitutes a stable part of the model. Consequently, this part should have more weight in the final vote because it has proved its reliability by the persistence of its matches. This persistence is interpreted as a factor $\omega$. At the initialization, we consider every SPiKeS from the initial bounding box equally reliable and set an initial weight $\omega^0 =1$ which is updated as
\begin{equation}\label{eq:omega}
\omega_i^{t+1} = (1-\beta)\omega_i^{t}+\beta \mathds{1}
\end{equation}
with $\mathds{1} =1$ if $S_i$ is a matching SPiKeS, $\mathds{1}=0$ otherwise. 

\begin{figure}[t]
\begin{center}
\includegraphics[width=1 \linewidth]{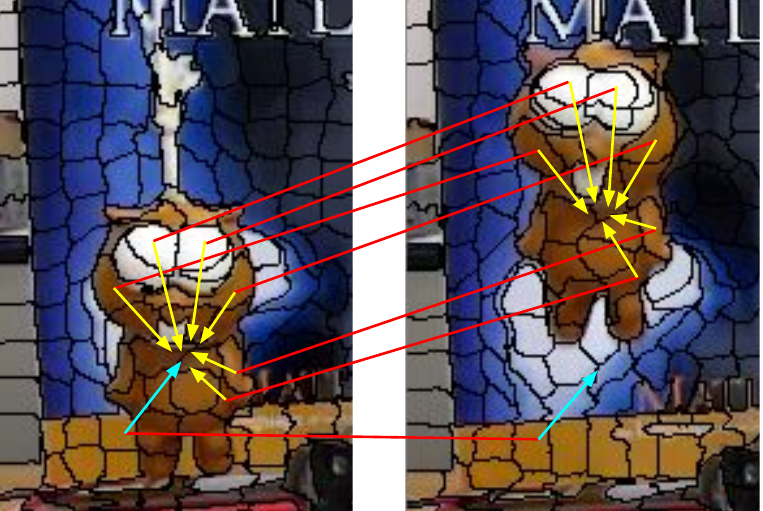}
\end{center}
\caption{A wrong vote of a background SPiKeS included in the model (cyan) will have a weak predictive factor $\phi$.}
\label{im:phi_match}
\end{figure}

However, suppose an unexpected part of the background is included in our model. It could match as often as a foreground part if it is also present in the other frames. In that case, the persistence factors $\omega$ would be the same while the foreground part should have more importance. It can be observed on figure \ref{im:phi_match} that, as a background SPiKeS will not follow the target, the center estimated by its vote will be far from the predicted location, whereas the foreground SPiKeS votes will be closer. To leverage this behaviour, we introduce a predictive factor $\phi$. Given a factor $\phi^0_l = 1$ at time $t=0$ for a SPiKeS $S_l^m$ belonging to the model, the predictive factor $\phi_l$ is updated such that it increases if the local prediction $\mathbf{x}_l$ given by the vote is near the final location:
\begin{equation}
\phi_l^{t+1} = \phi_l^{t}+\exp(-||\mathbf{x}_l-\mathbf{x}_t^T||^2)
\end{equation}
Those two factors $\omega$ and $\phi$ allow a SPiKeS to be more reliable if it often matches and votes correctly.

\begin{figure}[t]
\begin{center}
\includegraphics[width=1 \linewidth]{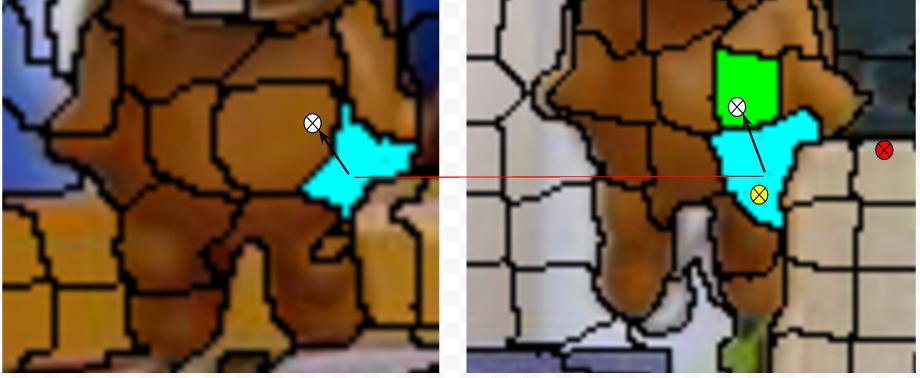}
\end{center}
\caption{A new keypoint (yellow, right) inside a matching superpixel (cyan) can be added to $\mathbf{K^f}$ because this superpixel belongs to the target, unlike the red keypoint. In the same way, a new superpixel (green, right) can be added to $\mathbf{S}^m$ because it includes a matching keypoint (white).}
\label{im:updateSpikes}
\end{figure}

{\textbf{Step 2 :}} \textit{SPiKeS insertion.}
To handle appearance changes such as pose change resulting in new parts not visible in the initial bounding box, one needs to add these new parts to the model. The main problem at this step is to avoid adding background parts that make the model drift. Therefore, instead of naively adding all the SPiKeS from the bounding box, we select only superpixels and keypoints that will make good SPiKeS candidates. Figure \ref{im:updateSpikes} illustrates how superpixels and keypoints help each other selecting good candidates. As the introduction stated, a keypoint inside a matching SPiKeS is assumed to belong to the target because of the boundary evidence given by the superpixel. Indeed, all the points inside that area are likely to belong to the same object. Similarly, a superpixel containing a matching foreground keypoint is more likely to belong to the target.

Let $\mathbf{s}_c$ and $\mathbf{k}_c$ be the sets of superpixels and keypoints candidates that meet these conditions. 
At first, the set $\mathbf{k}_c$ is added to the foreground keypoints pool $\mathbf{K^f}$. Then, the old SPiKeS from $\mathbf{S}^m$ refresh their structure with those new keypoints. Finally, we add the new SPiKeS made from the superpixels in $\mathbf{s}_c$ and the updated $\mathbf{K^f}$.

In order to complete the keypoint-based background model, keypoints detected around the estimated bounding box are added to $\mathbf{K^b}$ if they did not match the background keypoints.


{\textbf{Step 3 :}} \textit{SPiKeS deletion.}
As we add SPiKeS, our model grows and increases the complexity of the matching process. Furthermore, some SPiKeS may be irrelevant like redundant or background SPiKeS, which need to be deleted.  To keep a reasonable number of SPiKeS in our model, once a maximum size $N_m^{max}$ is exceeded, the $(N_m-N_m^{max})$ weakest SPiKeS are removed based on their persistence factor $\omega$. The same discarding method apply for $\mathbf{K^f}$ and $\mathbf{K^b}$. Thus a persistence factor $\omega^k$ is assigned to each keypoint, updated similarly to equation \ref{eq:omega}, so that the weakest ones can be identified.

\begin{figure*}[]
     \centering
     \subfloat[][Overall]{\includegraphics[width = 4.5cm]{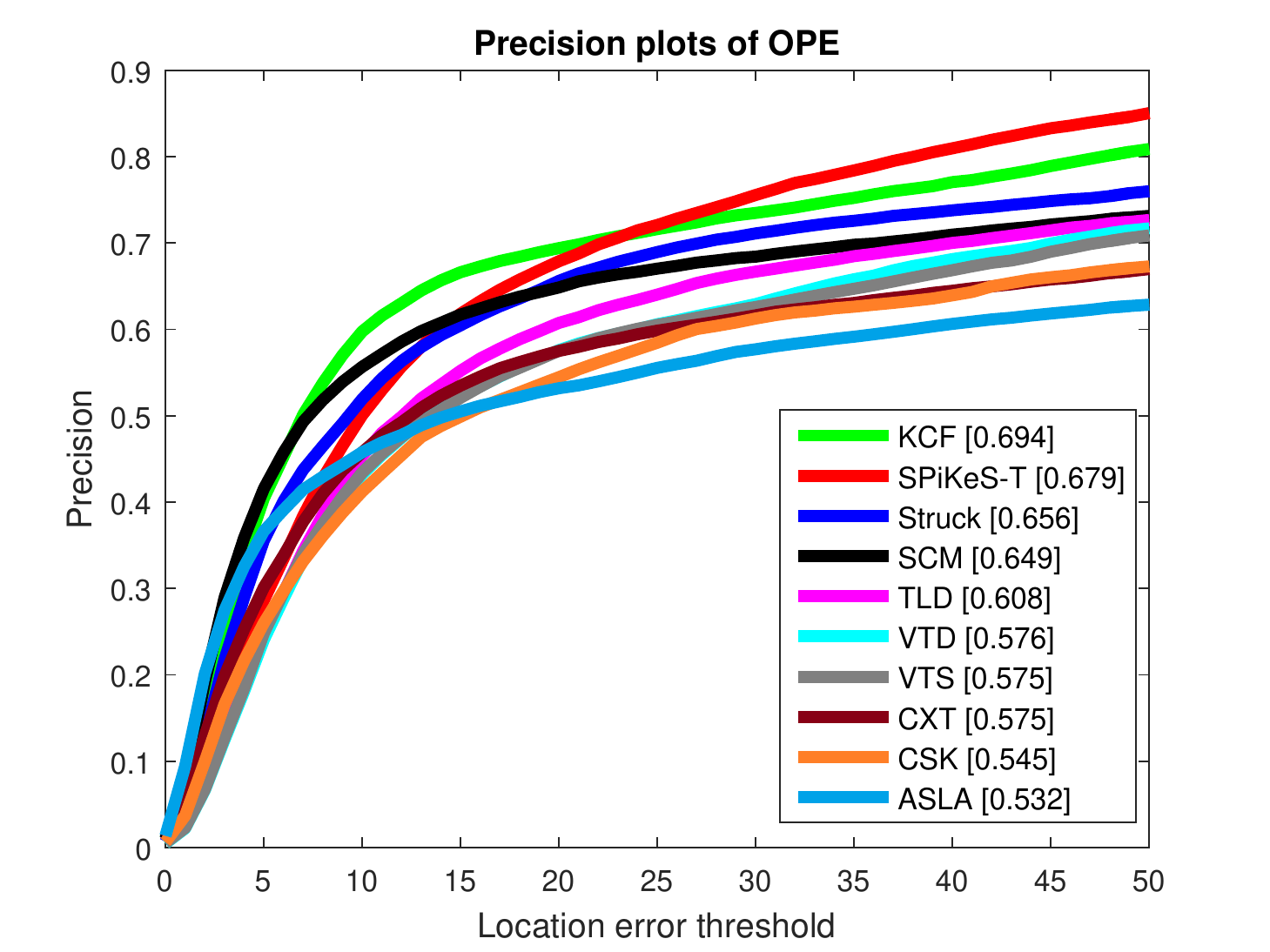} \includegraphics[width = 4.5cm]{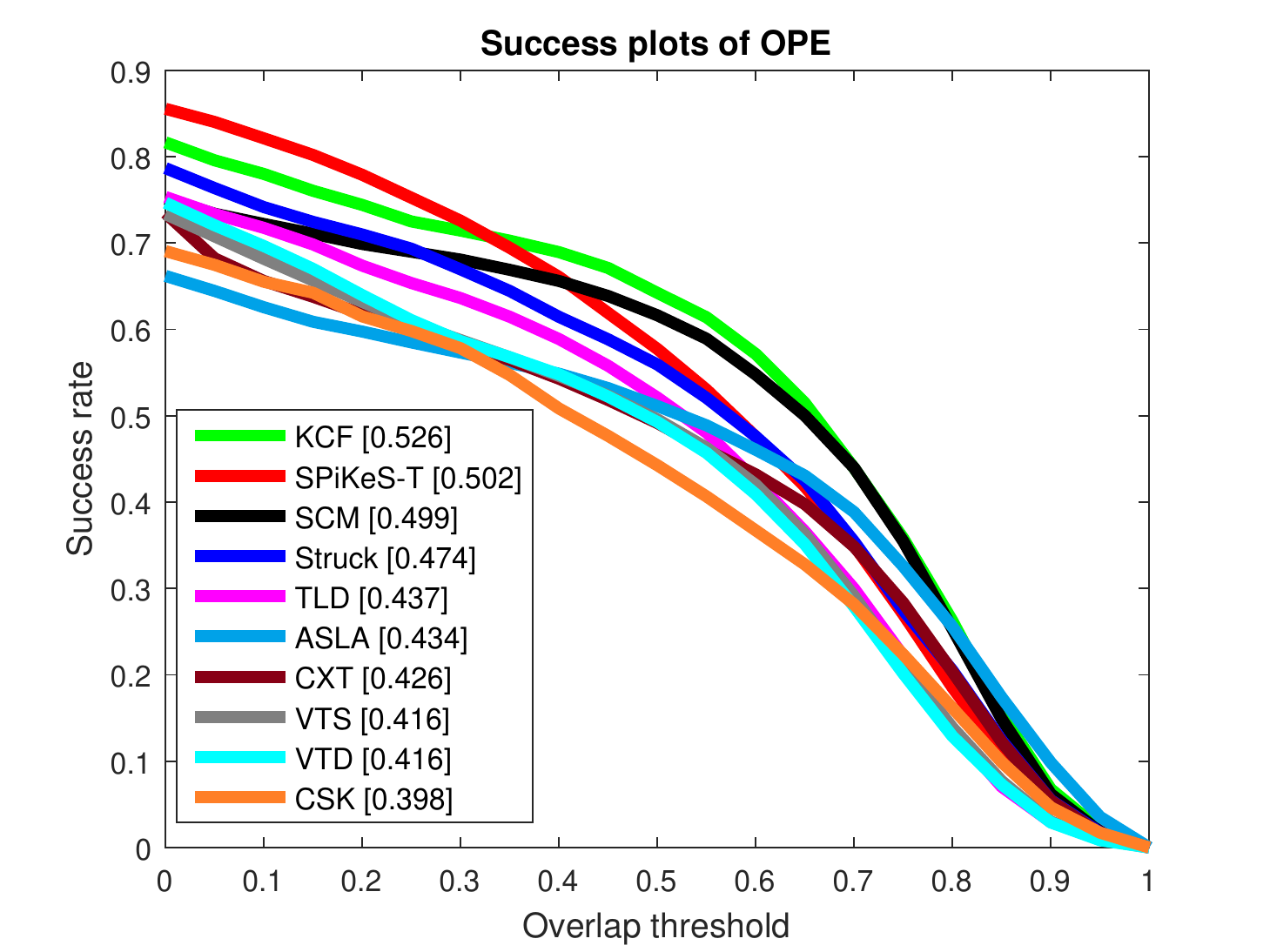}\label{im:Overall}}
     \subfloat[][Background Clutter]{\includegraphics[width = 4.5cm]{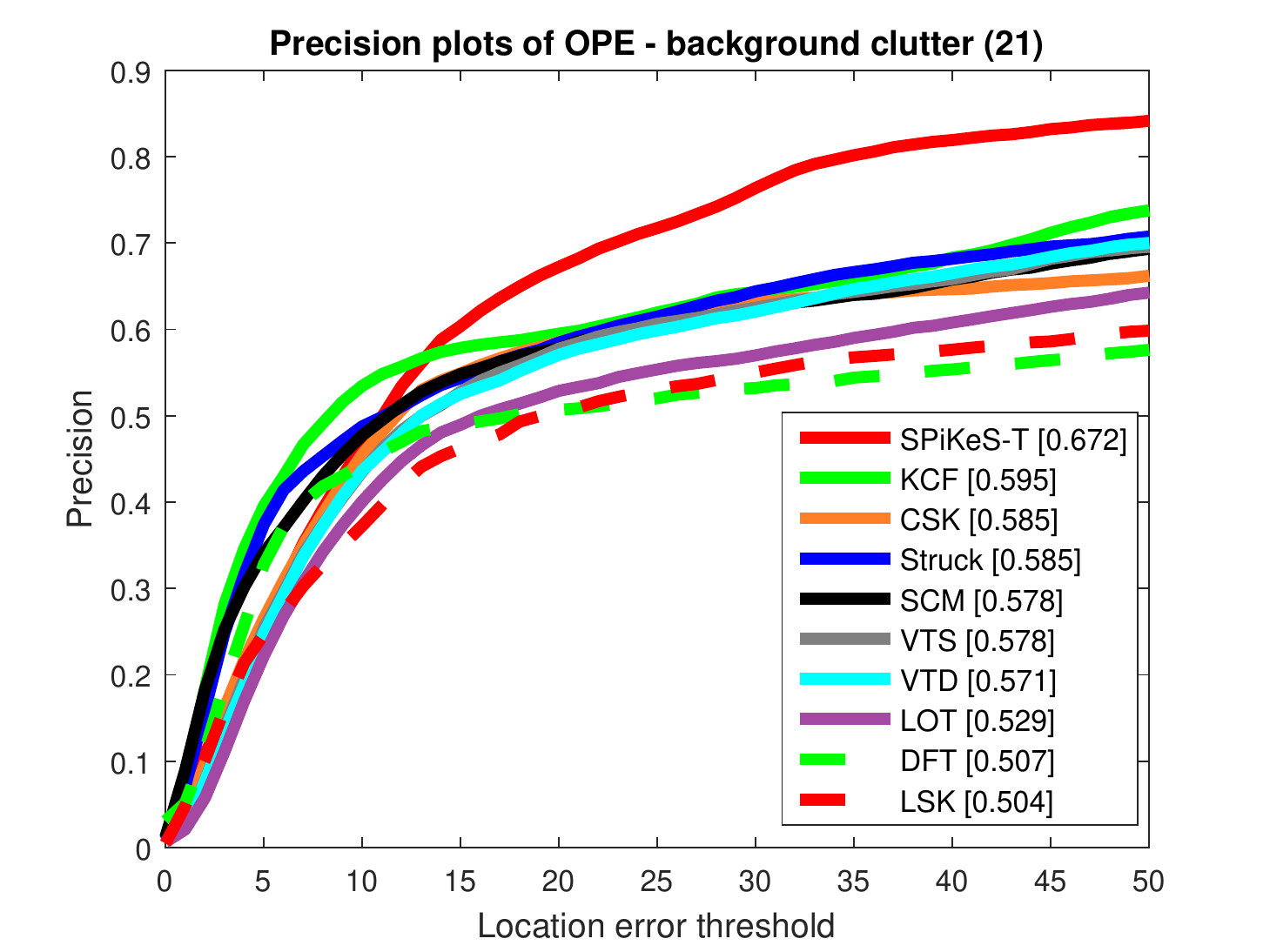} \includegraphics[width = 4.5cm]{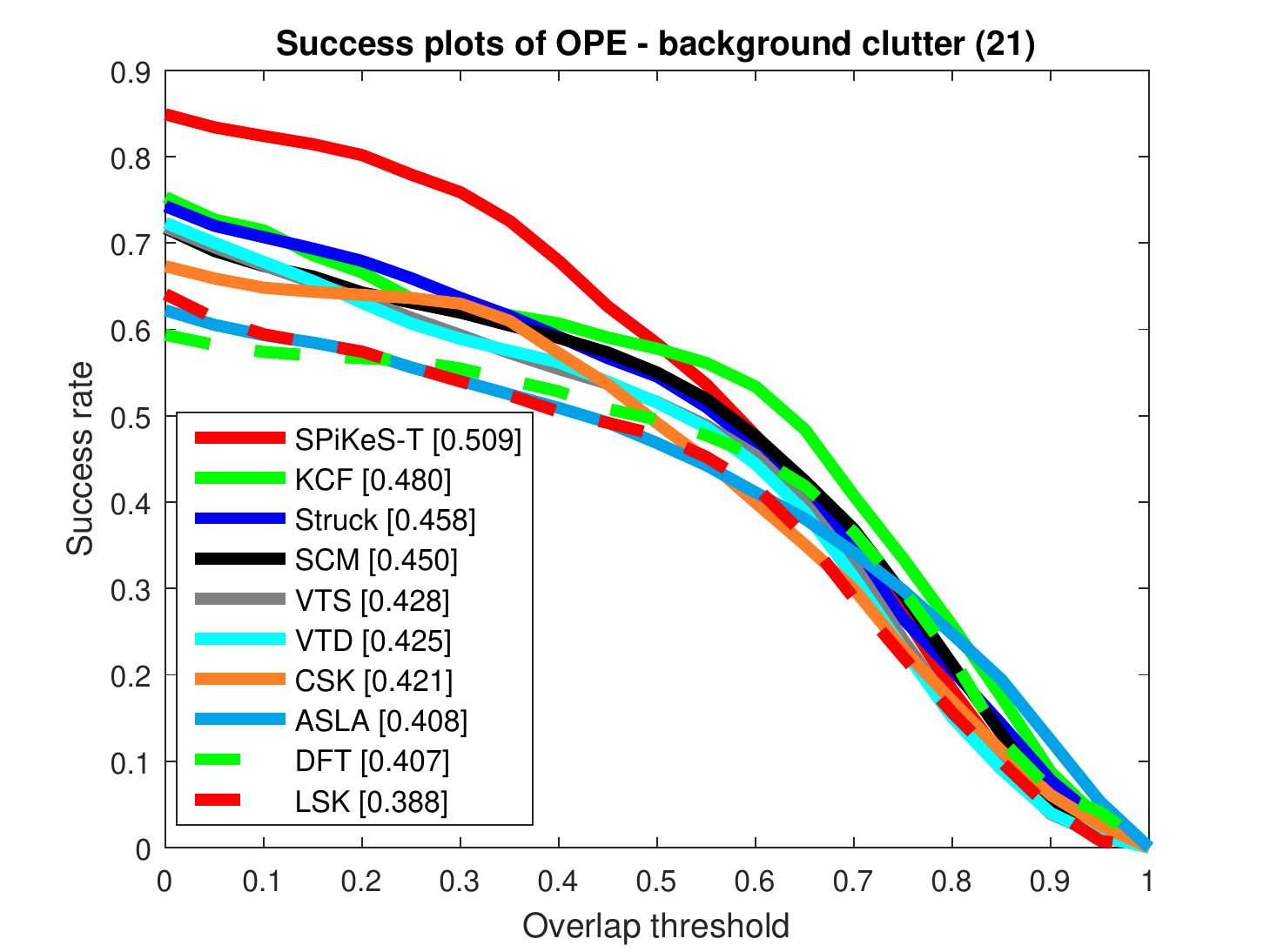}\label{im:bc}}\\
      \subfloat[][Deformation]{\includegraphics[width = 4.5cm]{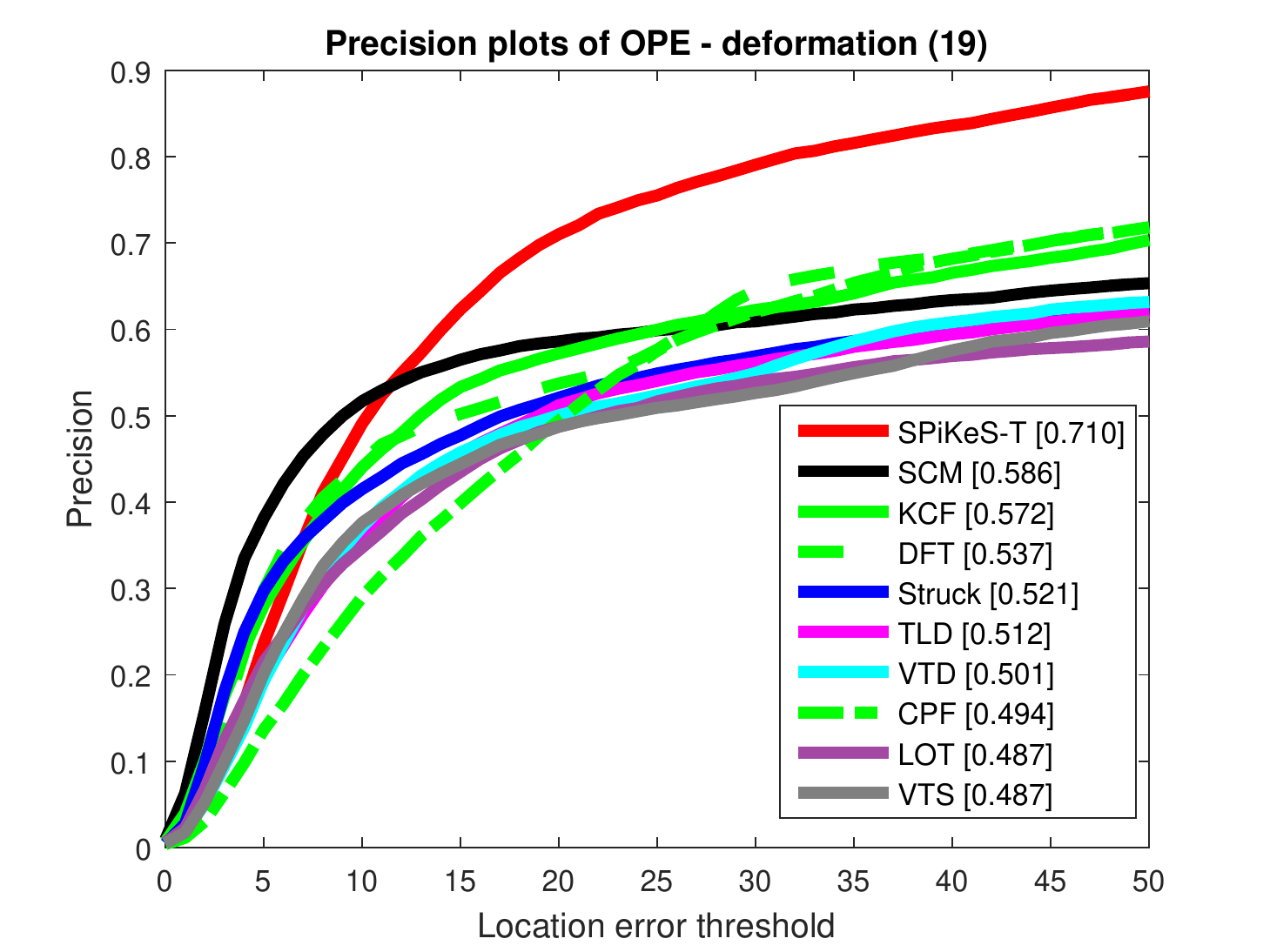} \includegraphics[width = 4.5cm]{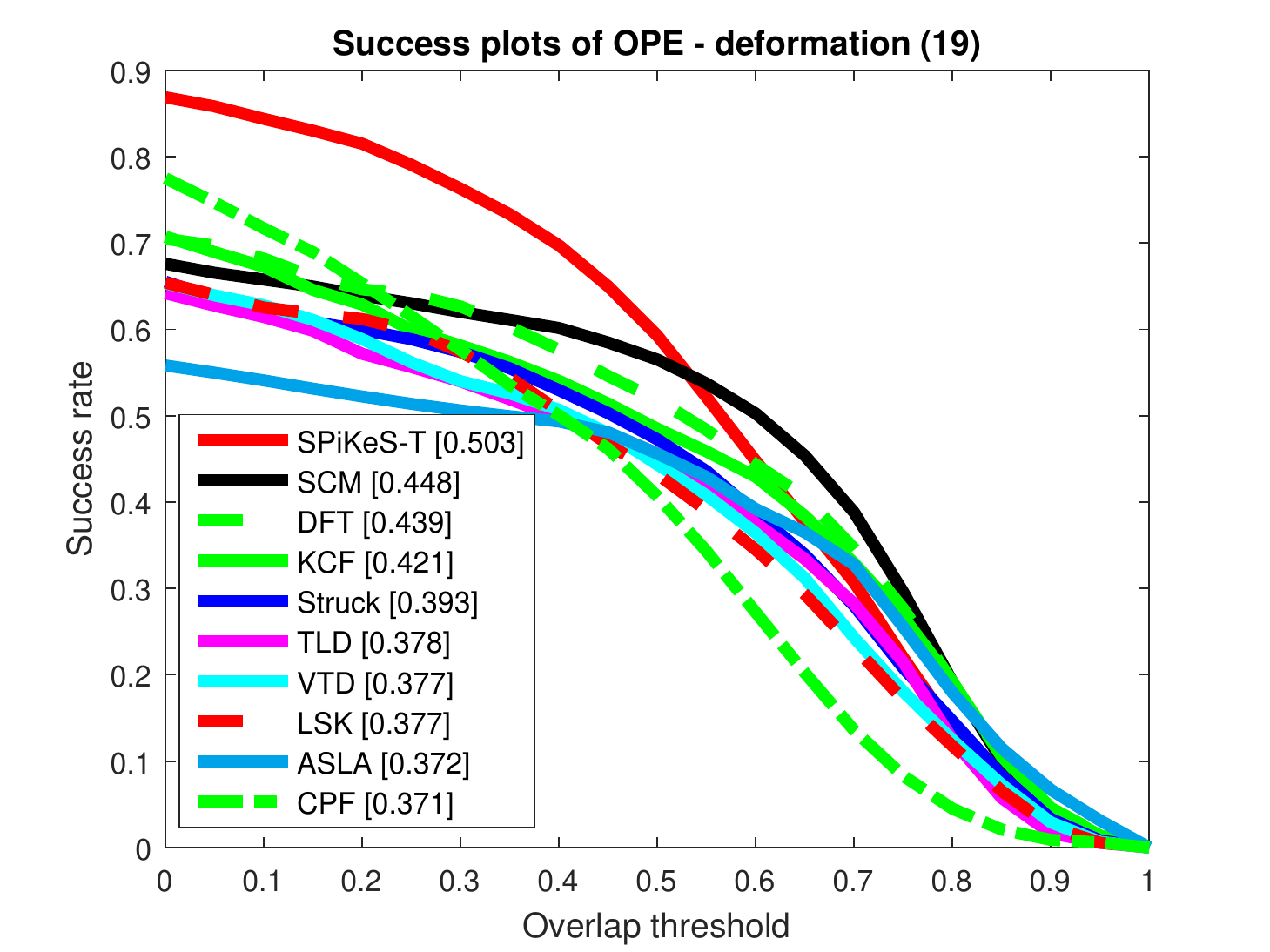}\label{im:def}}
     \subfloat[][Motion Blur]{\includegraphics[width = 4.5cm]{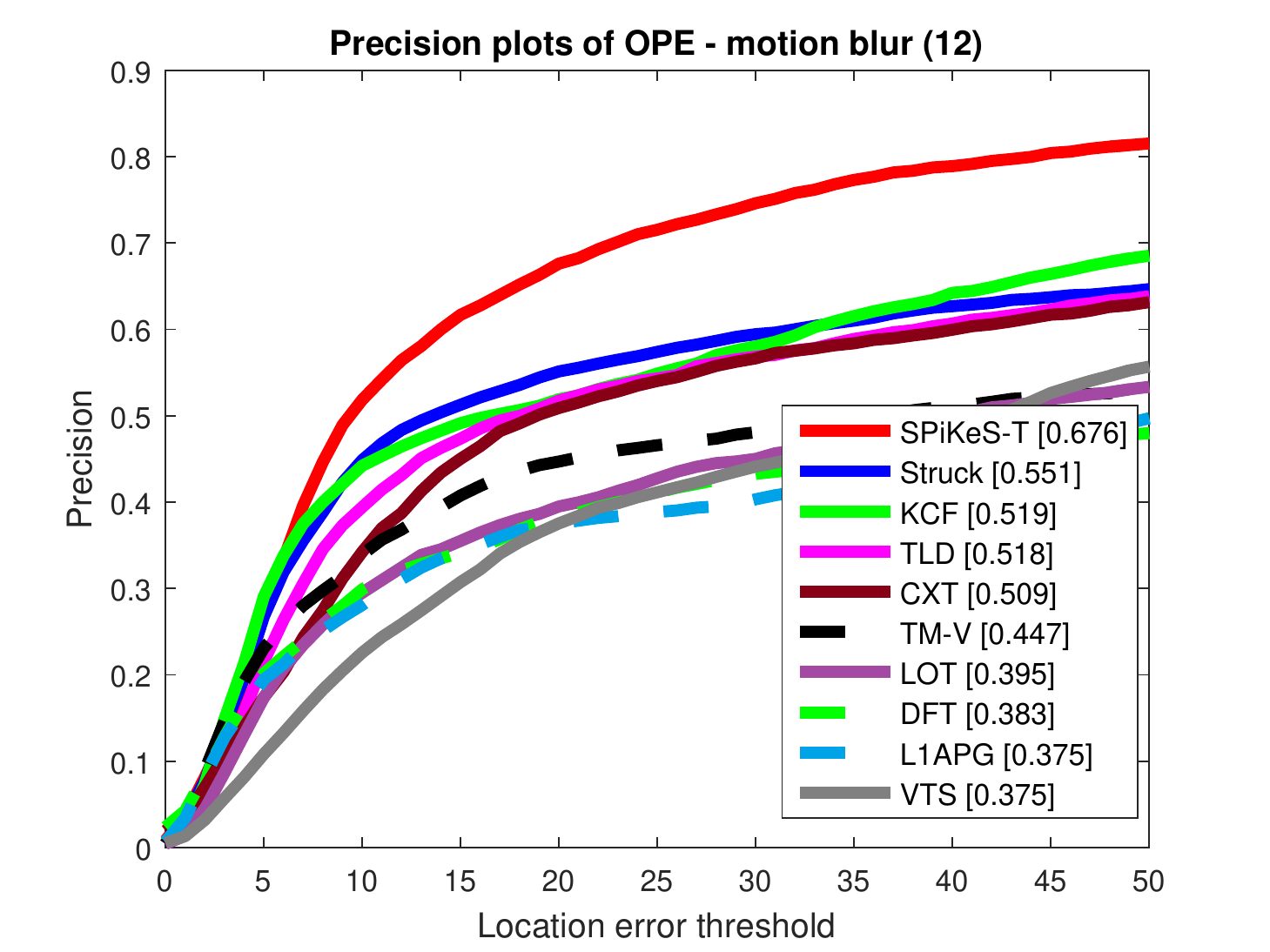} \includegraphics[width = 4.5cm]{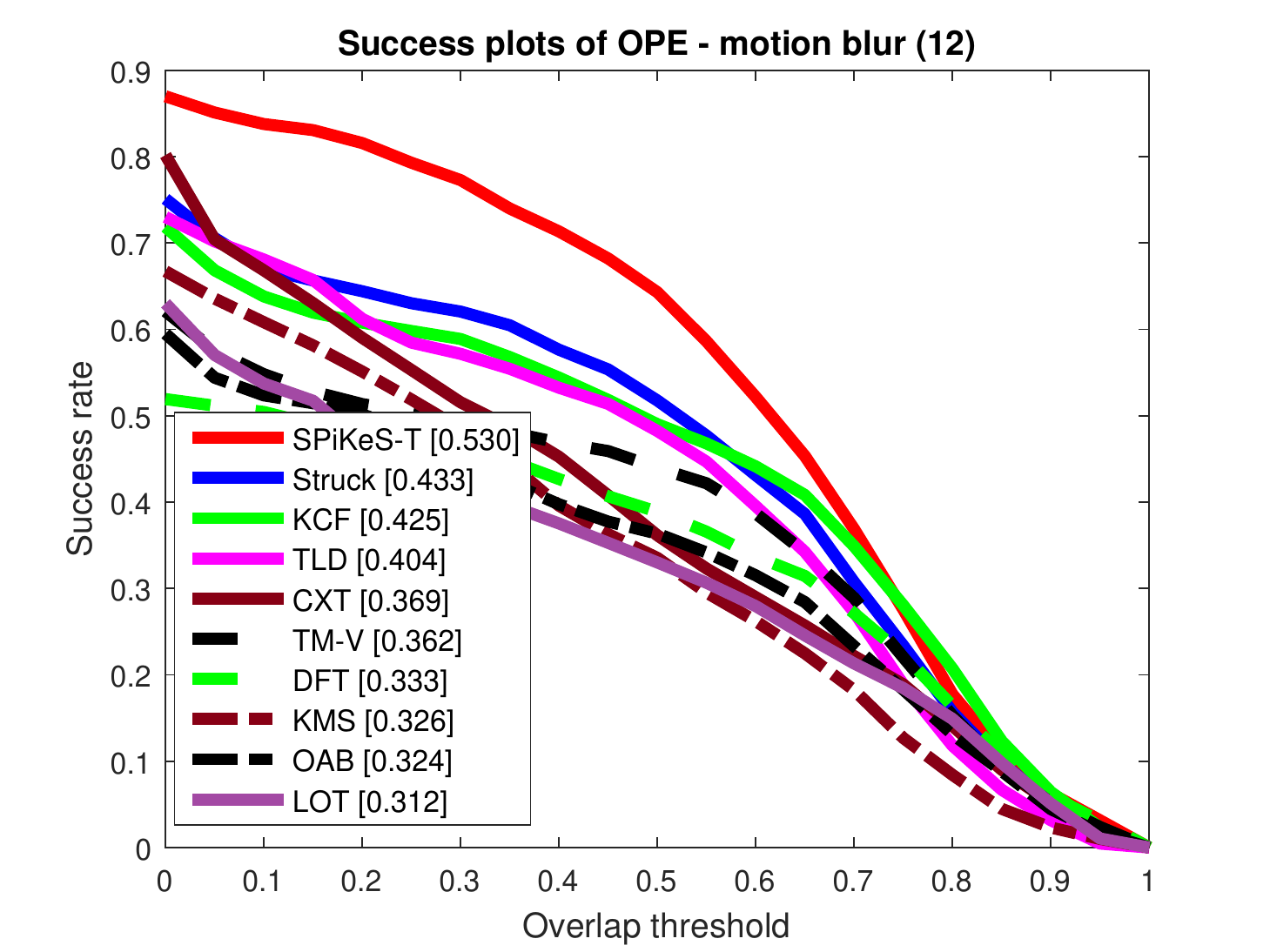}}\\
          \subfloat[][In-plane Rotation]{\includegraphics[width = 4.5cm]{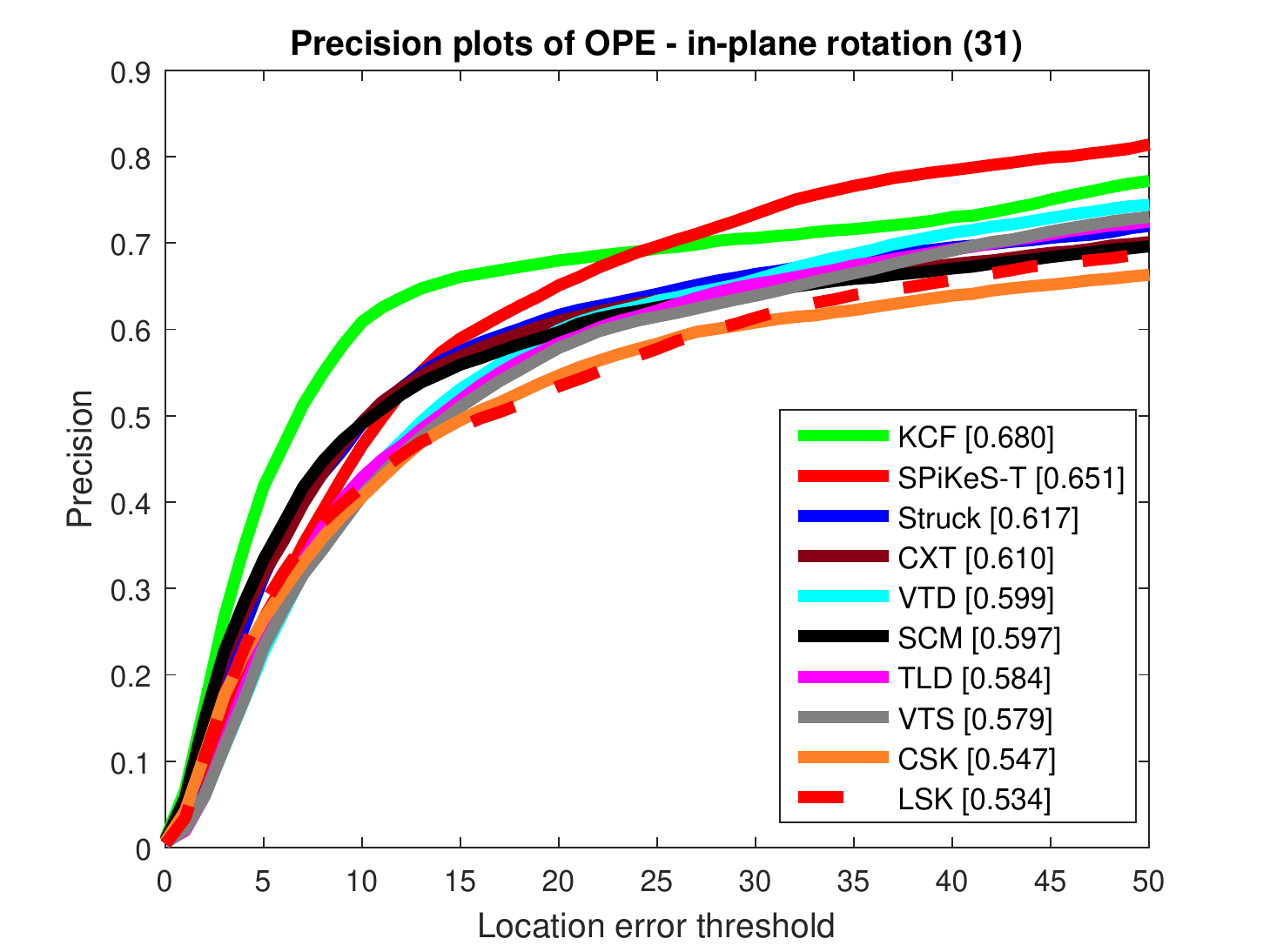} \includegraphics[width = 4.5cm]{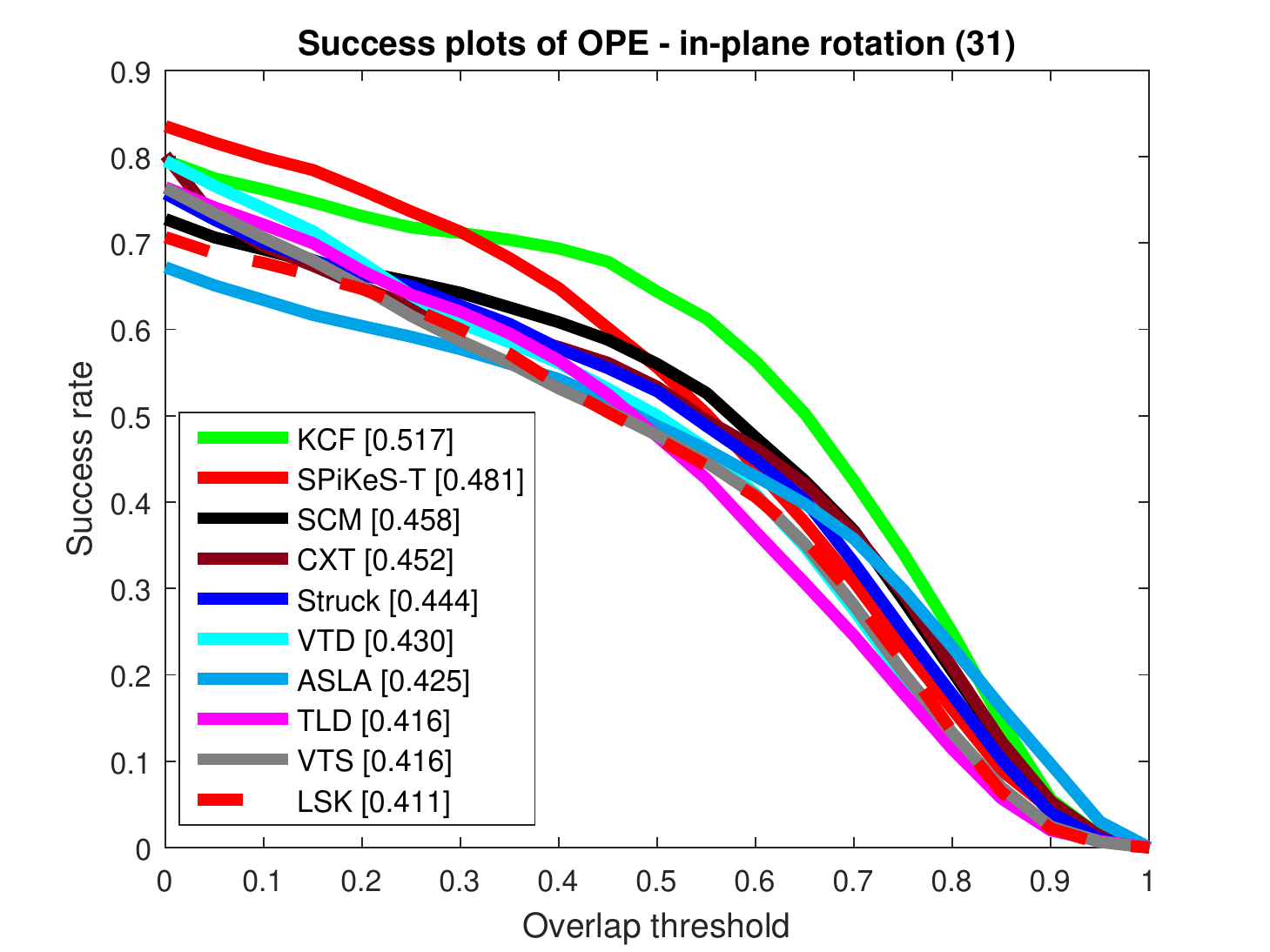}}
     \subfloat[][Illumination Variation]{\includegraphics[width = 4.5cm]{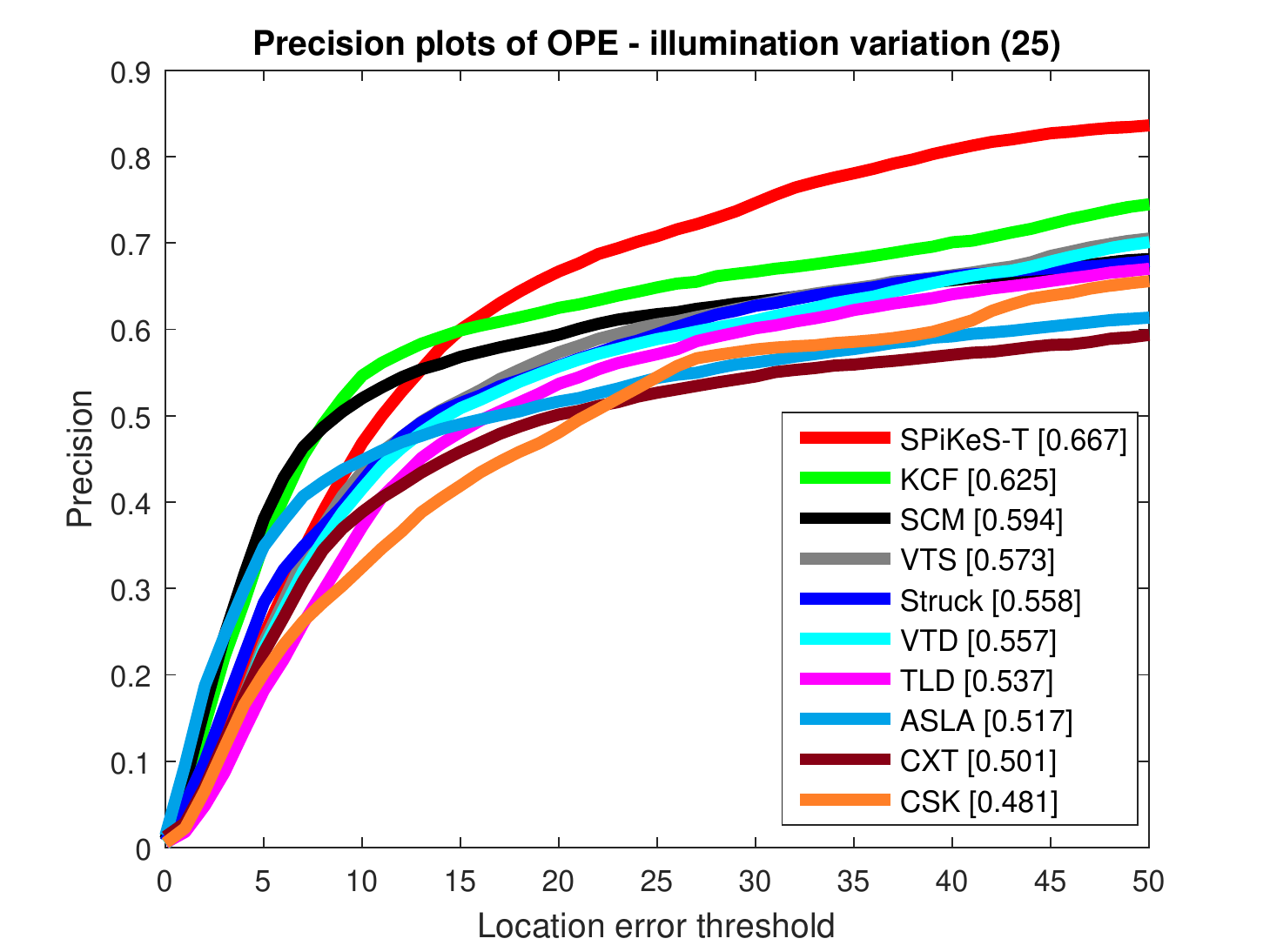} \includegraphics[width = 4.5cm]{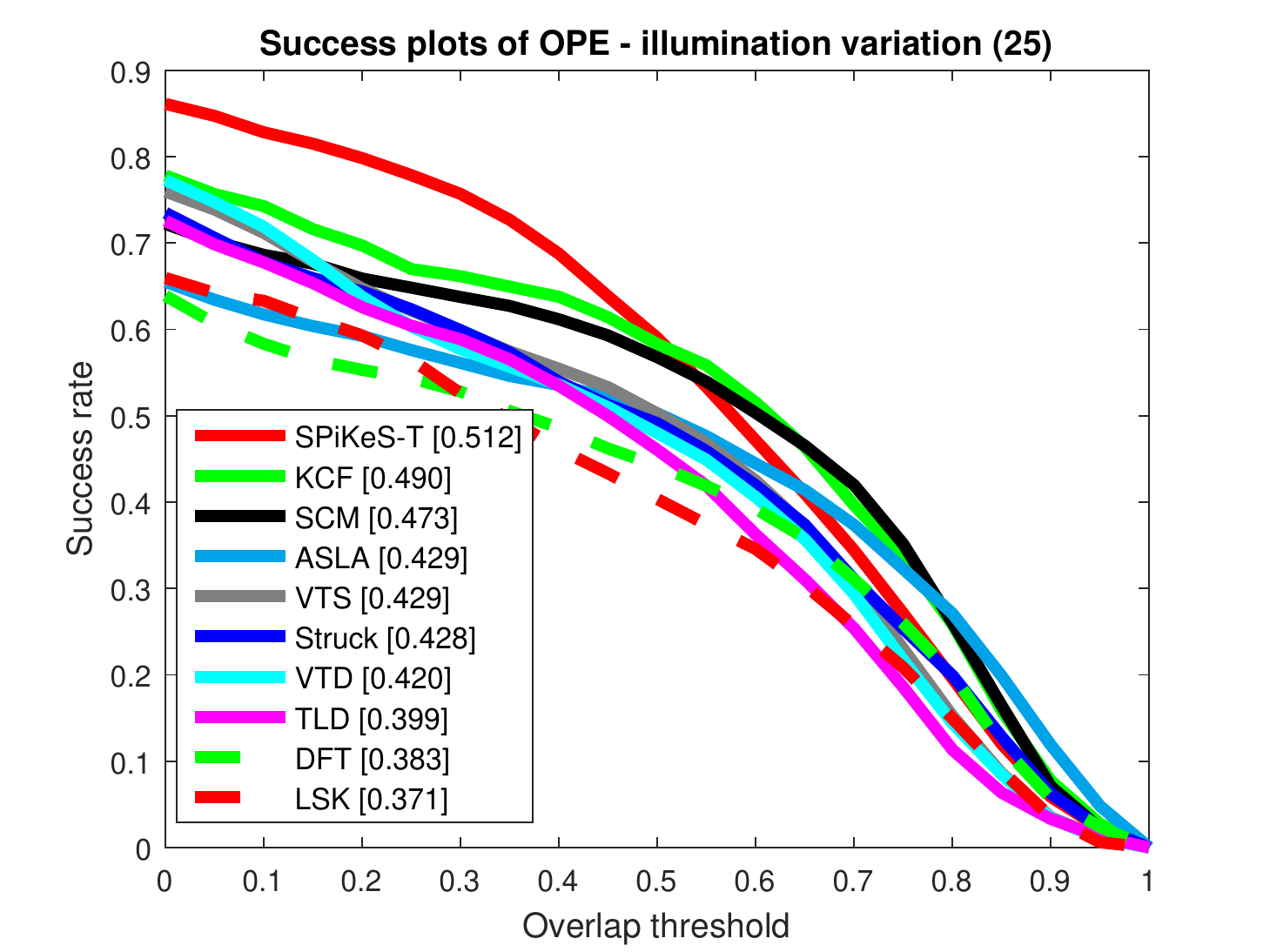}\label{im:iv}}\\
     \subfloat[][Scale Variation]{\includegraphics[width = 4.5cm]{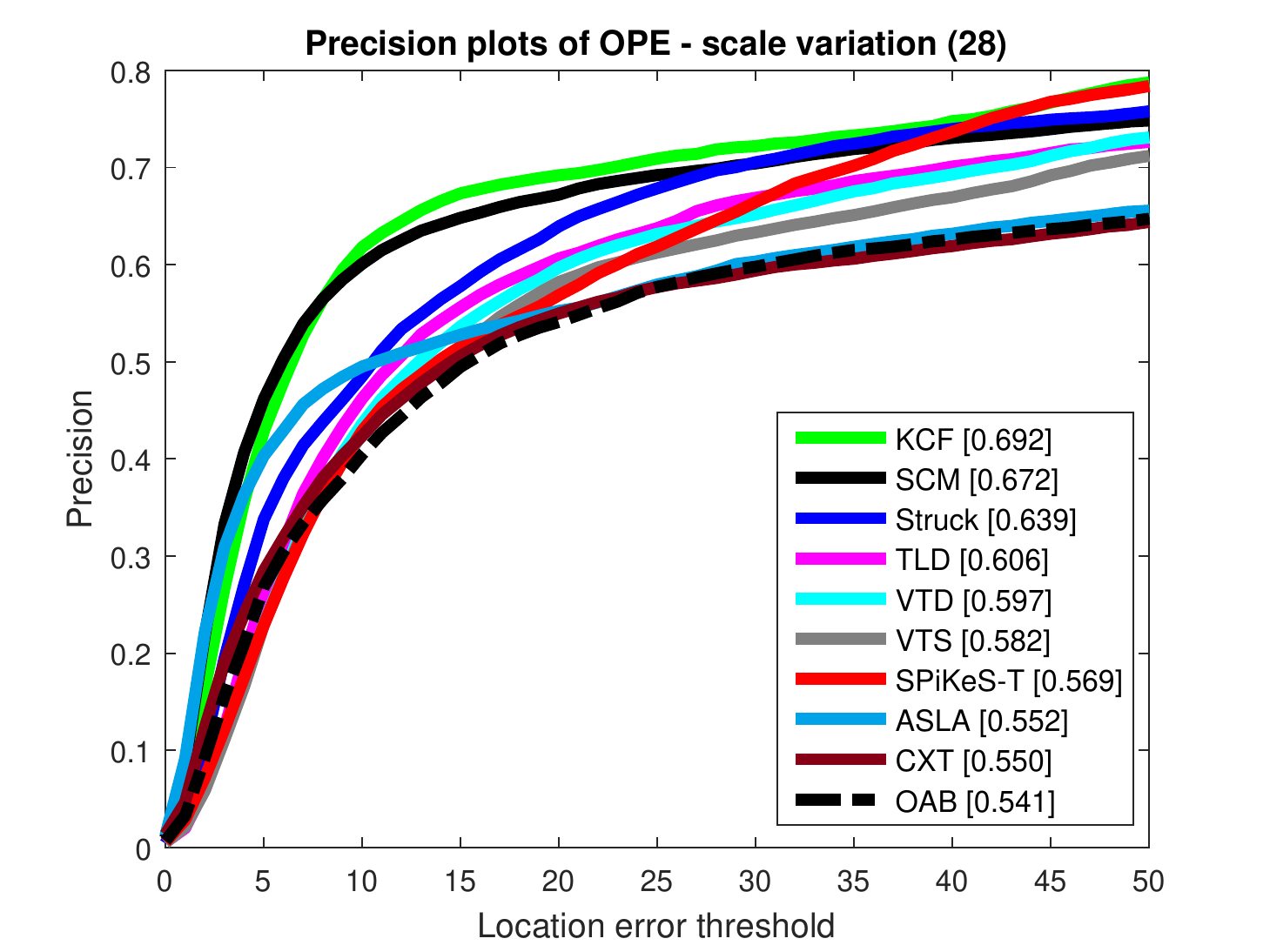} \includegraphics[width = 4.5cm]{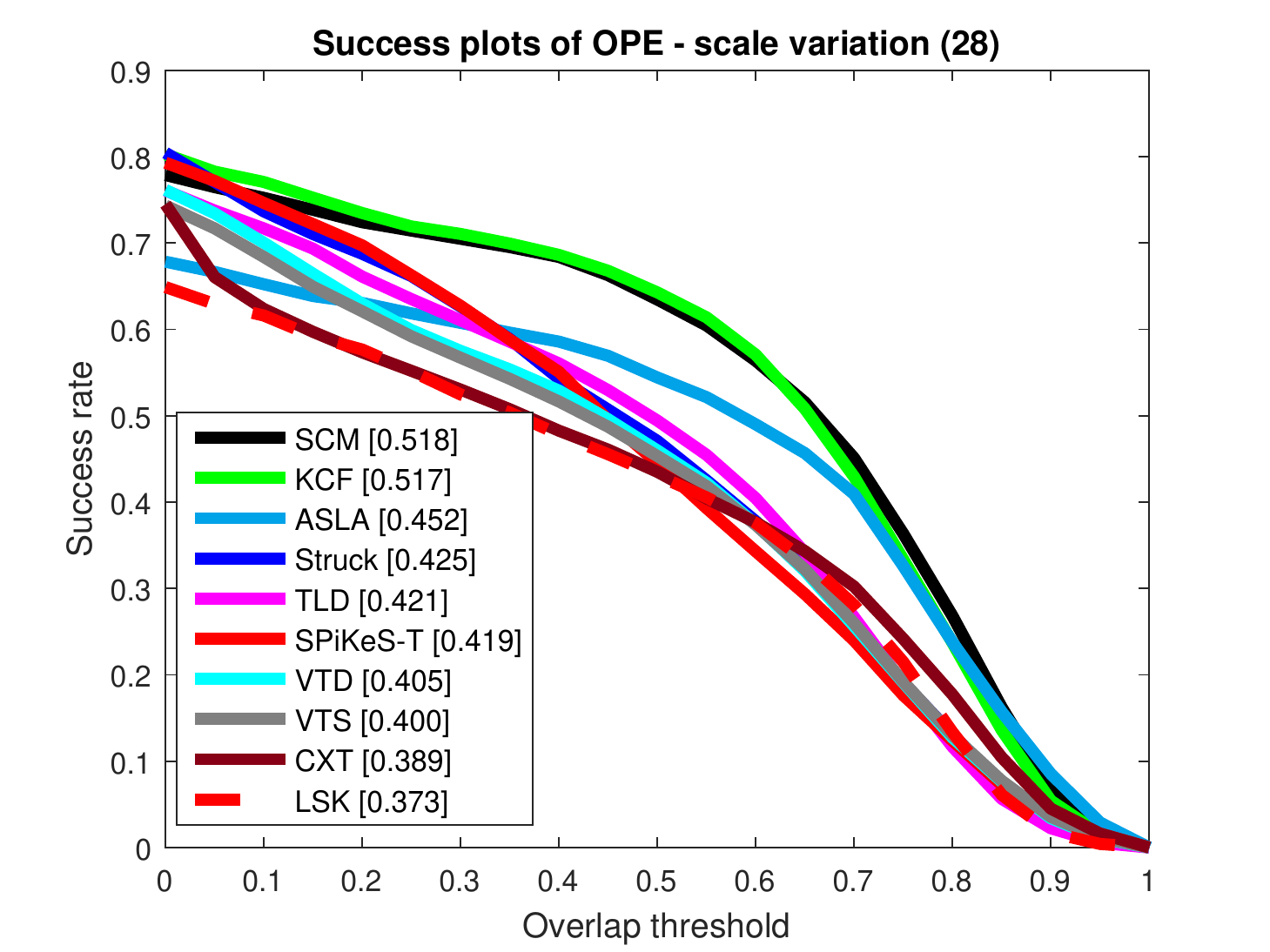}\label{im:sv}}
     \subfloat[][Fast Motion]{\includegraphics[width = 4.5cm]{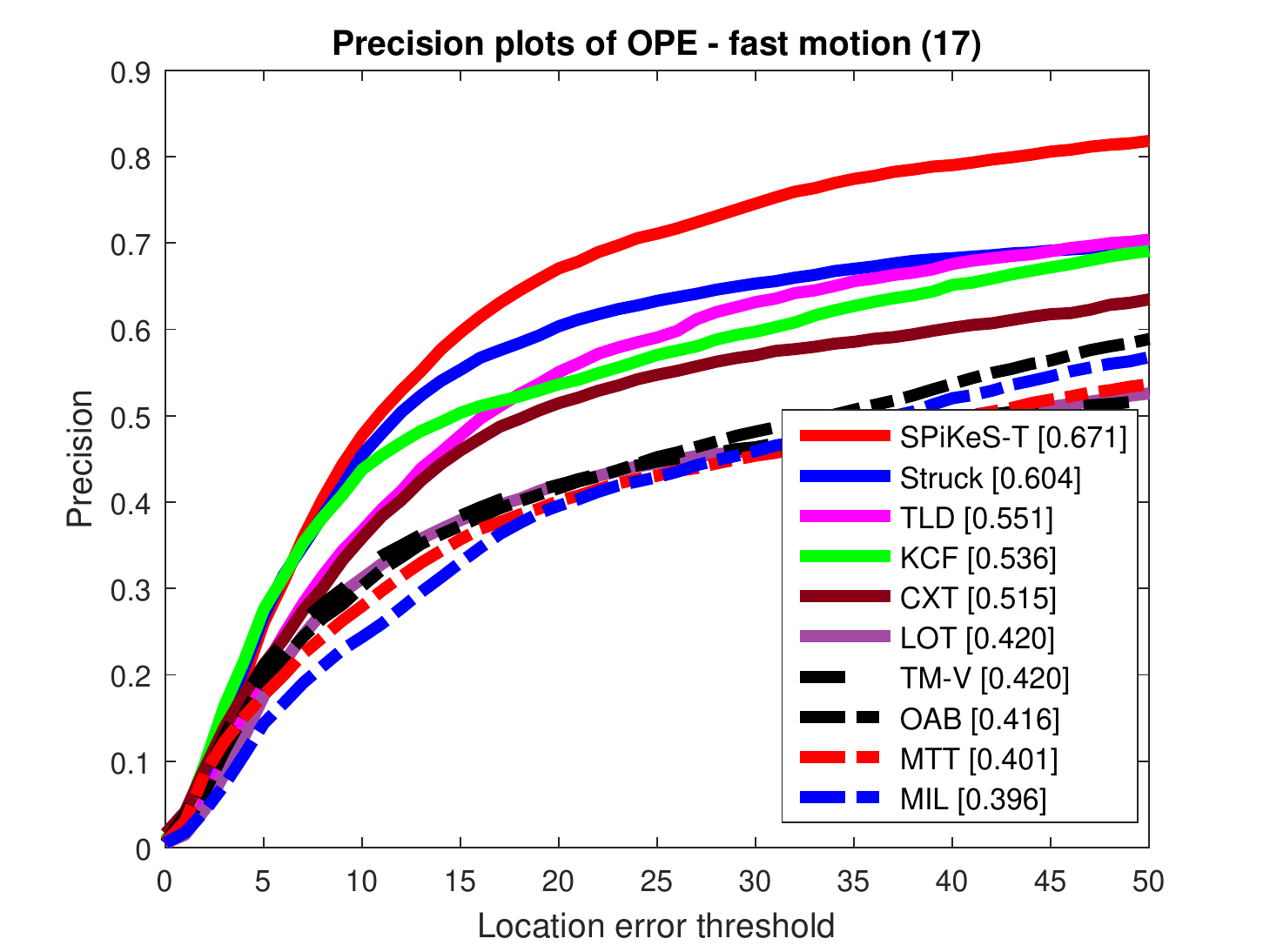} \includegraphics[width = 4.5cm]{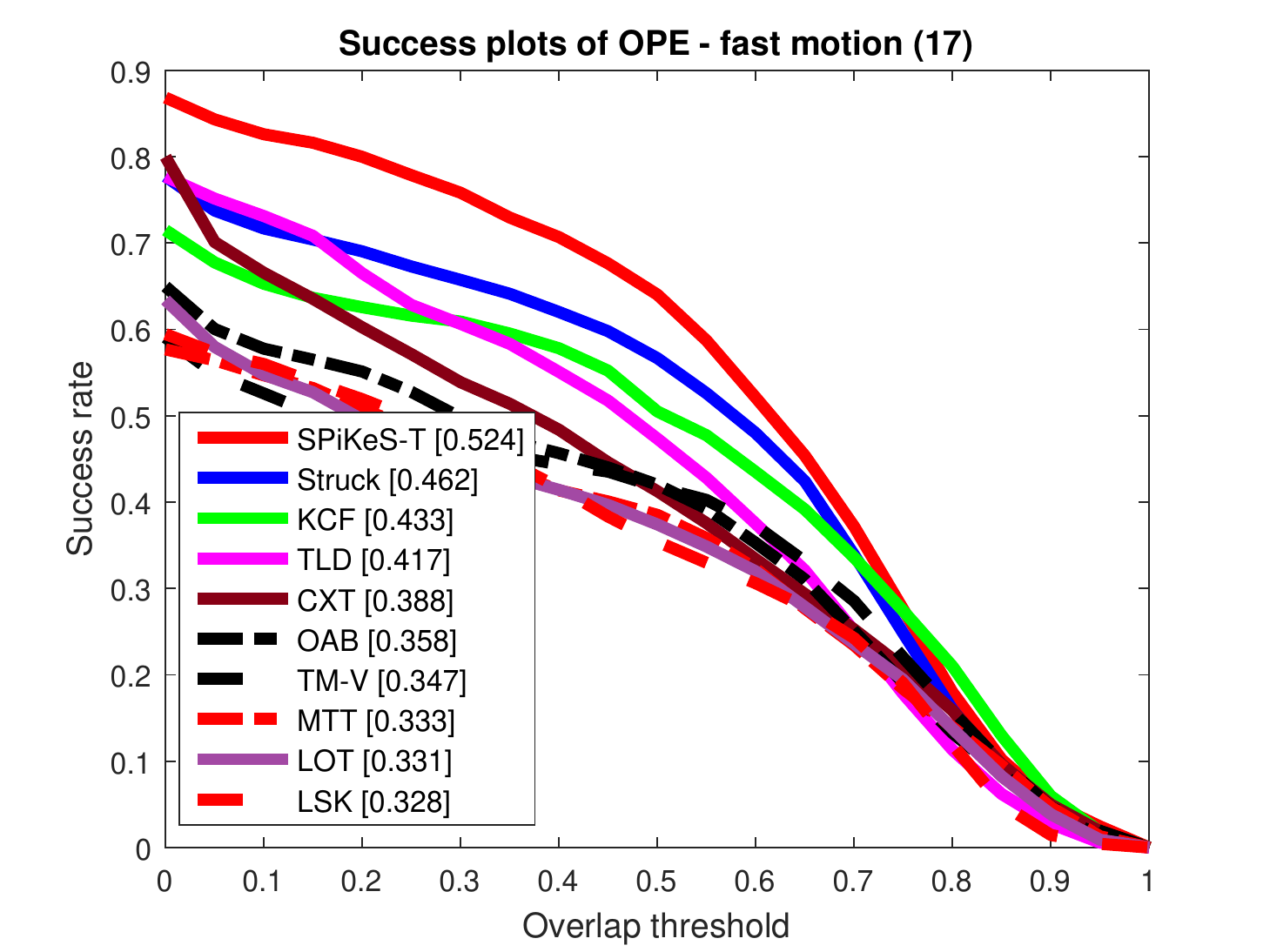}\label{im:fm}}\\
      \subfloat[][Out-of-plane Rotation]{\includegraphics[width = 4.5cm]{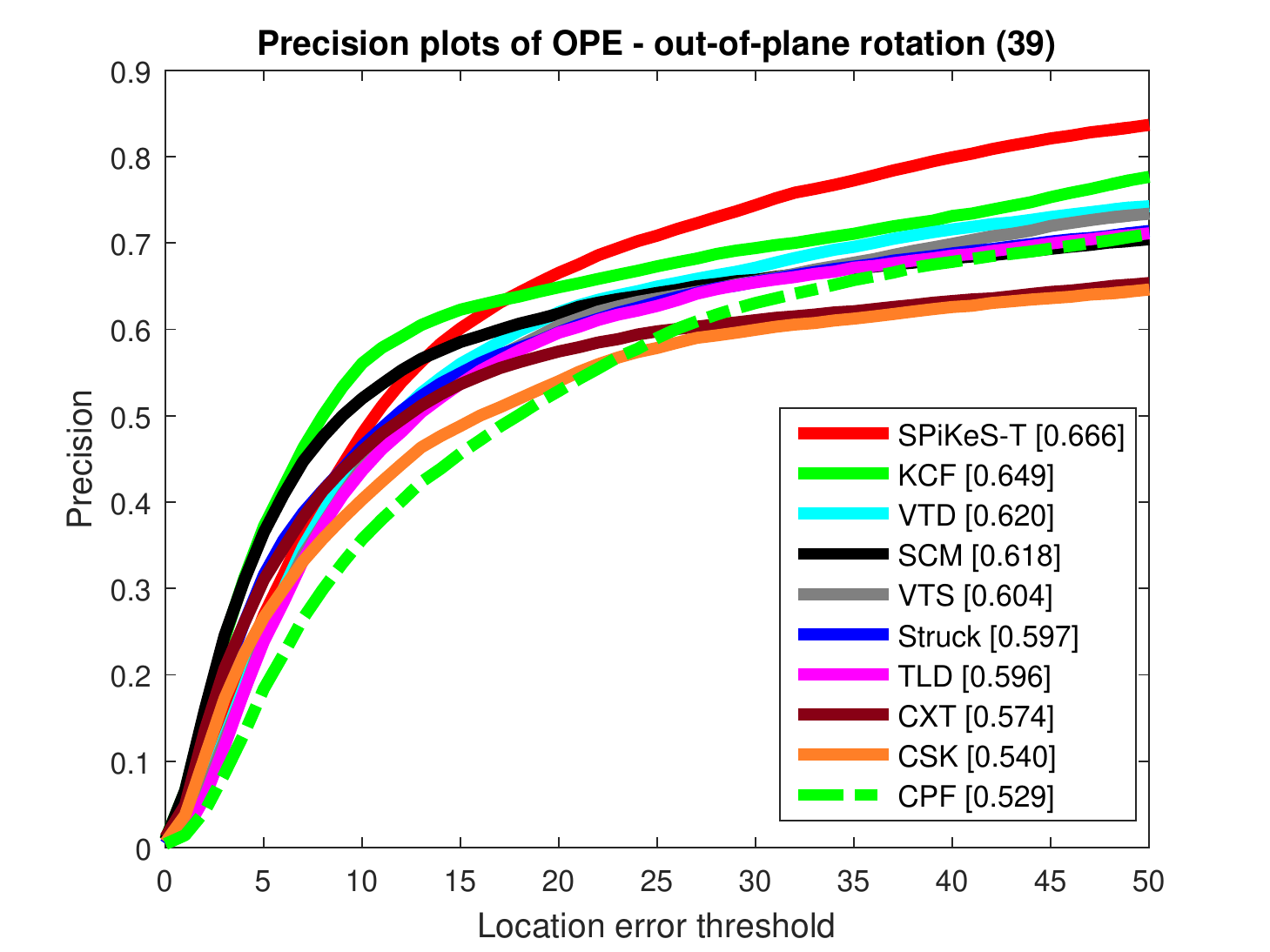} \includegraphics[width = 4.5cm]{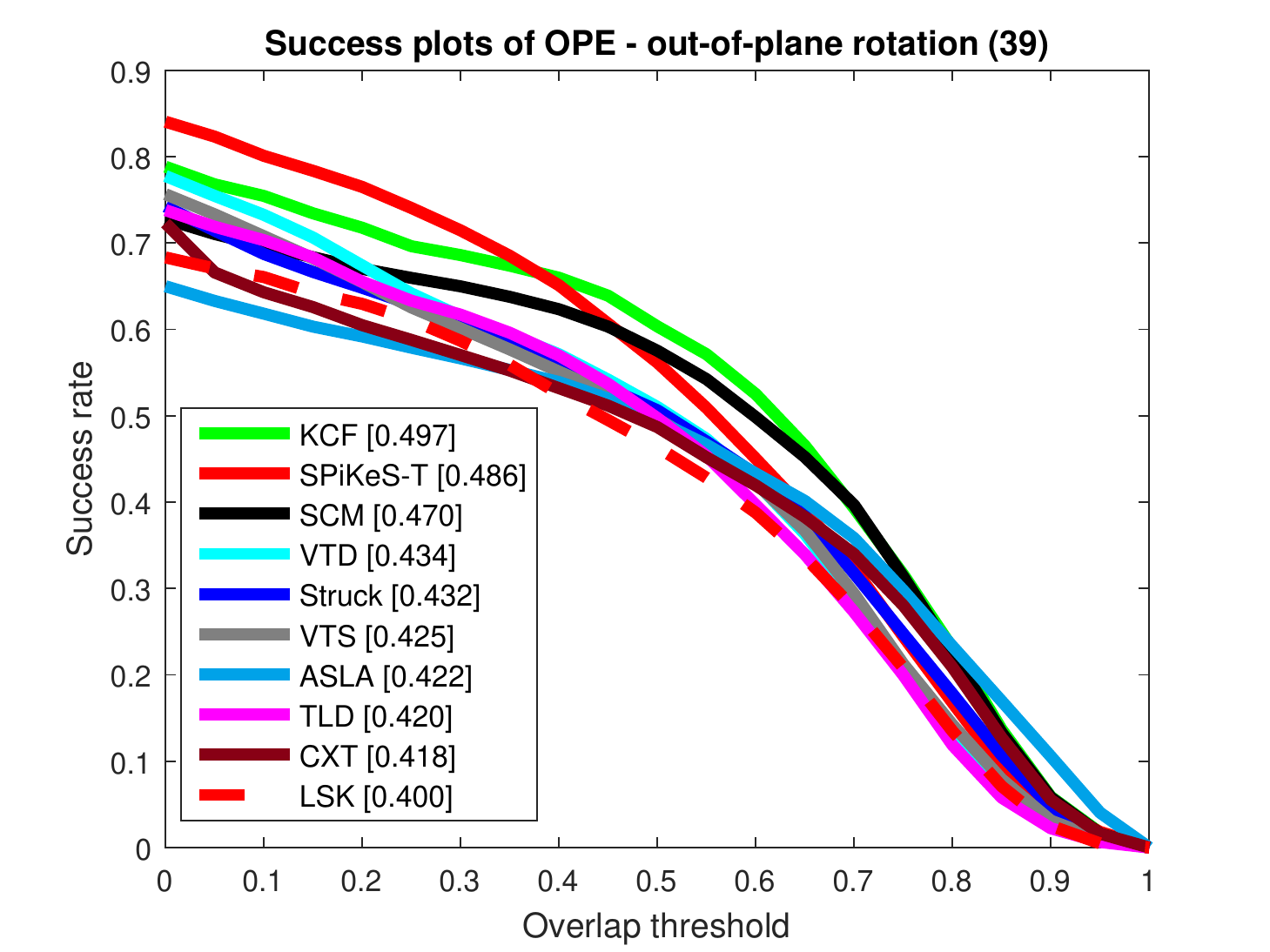}}
     \subfloat[][Occlusion]{\includegraphics[width = 4.5cm]{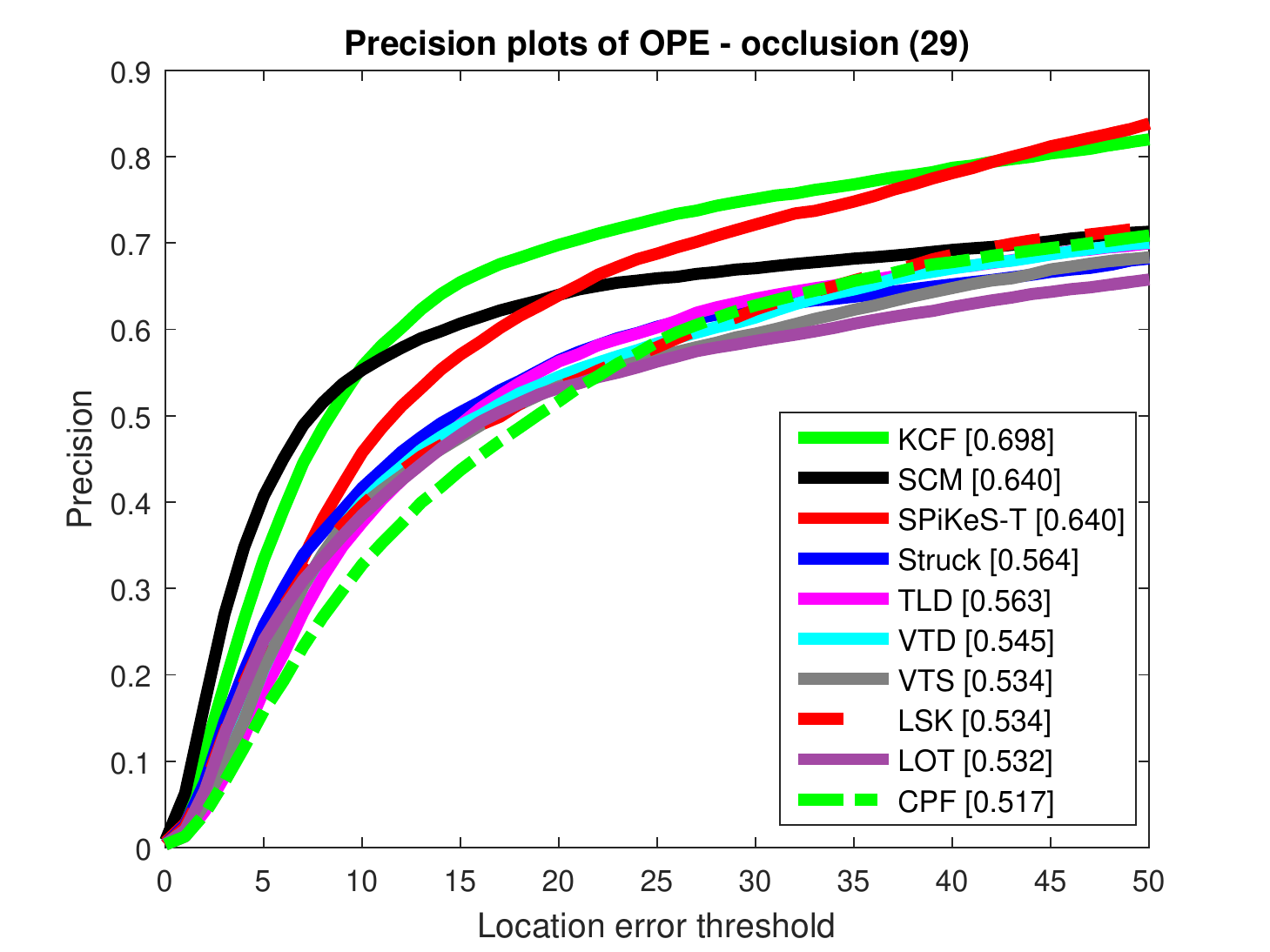} \includegraphics[width = 4.5cm]{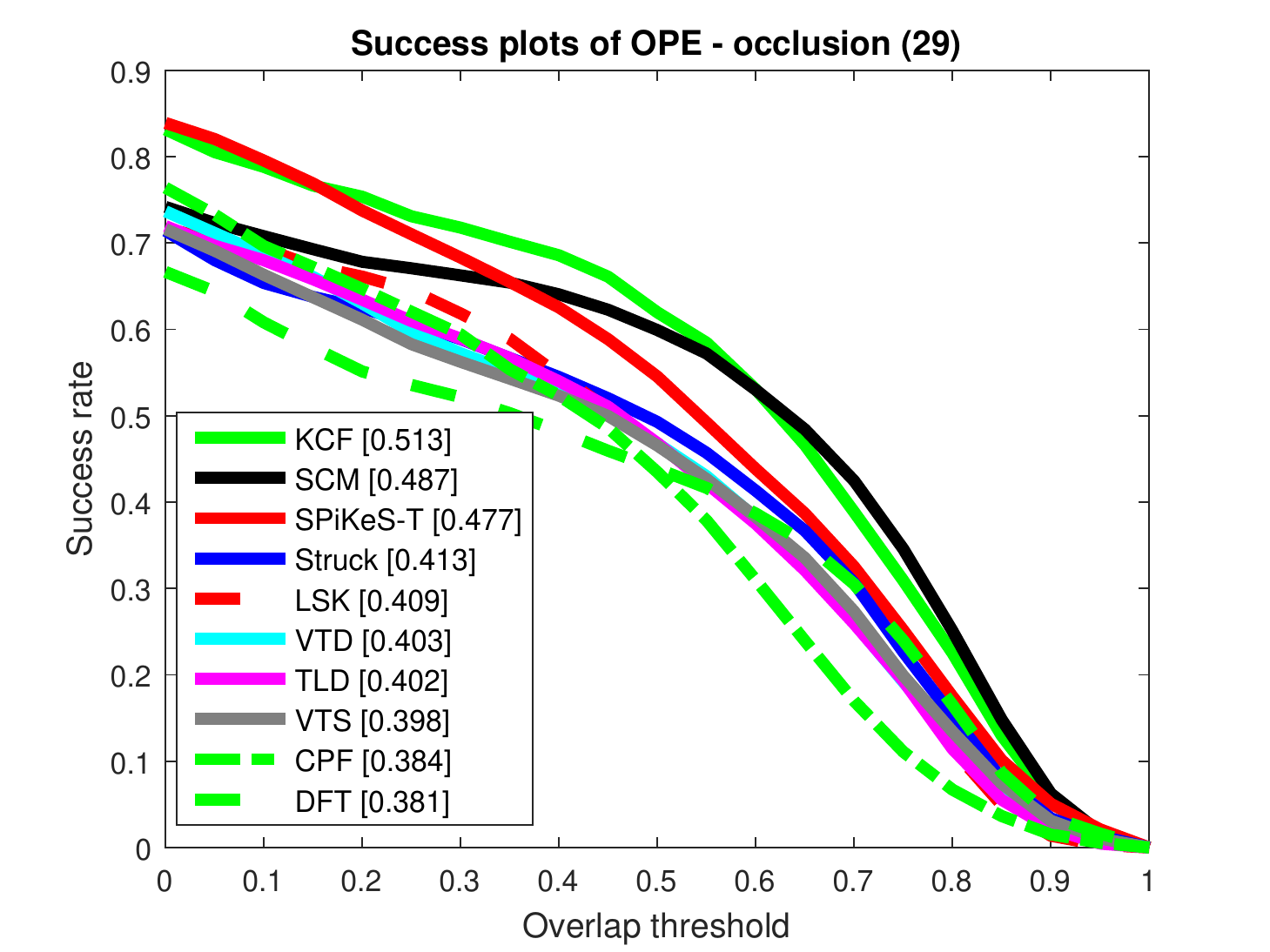}\label{im:occ}}
     \caption{Precision and Success plots for the one-pass evaluation (OPE) on OTTB. The number into brackets is the number of videos in the subset.}
     \label{im:result}
\end{figure*}

\section{Experiments}\label{sec:experiment}
In this section, we first present details of our implementation and the values for our parameters. Afterwards, we evaluate our method with the procedure proposed by \cite{WuLimYang13} and compare our results to state-of-the-art trackers.
\subsection{Experimental setup}
For the oversegmentation, we choose the SEEDS superpixels \cite{SEEDS} which have smooth boundaries and similar shapes in addition to be one of the fastest superpixel segmentation in the litterature. The size of a superpixel depends on the initial bounding box of dimension $w_B\times h_B$. In order to have about 30 superpixels included in the initial bounding box, a frame of dimension $w\times h$ should be segmented in $N=\frac{wh}{30 w_Bh_B}$ superpixels of diameter $D^s=\sqrt{\frac{wh}{N}}$. Their HSV color histogram is quantified in $6\times 6\times 6$ bins and normalized.
Similarly to \cite{multilevelSpx}, Grabcut \cite{grabCut} is used on the first frame to select foreground superpixels inside the given bounding box. This process makes the model more accurate as it avoids including background superpixels in the initial model.
As for the keypoints and their descriptors, we use the SIFT algorithm \cite{SIFT} which produces scale and rotation invariant keypoints robust against illumination variation. A match between keypoints is defined as proposed in \cite{SIFT} with a ratio threshold $\theta_{lo} = 0.75$. 
When building the SPiKeS, each superpixel searches its keypoints in a surrounding region of radius $R = 2 D^s$. We limit the size of $\mathbf{S}^m$ to 3 times the number of superpixels in the initial bounding box. The keypoints model $\mathbf{K^f}$ and $\mathbf{K^b}$ are limited to $1000$. 
During the matching process, the color threshold is set to $\theta_c = 0.7$, the score parameter to $\lambda_1=1$ and the motion constraint factor to $\lambda_2 = 4D^s$. At the update stage, the occlusion parameter is set to $\theta_o =3$. A smooth appearance adaptation is obtained with interpolation factor $\alpha_v=\alpha_f = 0.1$ and learning factor $\beta = 0.1$. Finally, when a new SPiKeS is added, it starts with a weak persistence factor $\omega_{min}=0.1$ such that it could be discarded quickly if it does not match directly in the next frame. 
Our results can be reproduced with our C++ implementation available online\footnote{\url{https://github.com/fderue/SPiKeS_T}}. The following evaluation gives an average of 3 frames per second on a 3.4 GHz CPU with 8 GB memory, without code optimization. Note that most of the time is spent on superpixels segmentation, keypoints computation and matching, which are tasks that could be implemented on GPU to improve execution speed.

\subsection{Evaluation}
\subsubsection{Comparison to the state-of-the-art}
The CVPR2013 Online Object  Tracking Benchmark (OTTB) of Wu et al.\cite{WuLimYang13} allows us to evaluate our approach against 29 state-of-the-art trackers over a dataset of 51 challenging sequences. The given groundtruth is a rectangular bounding box whose center corresponds to the target location. We also added a more recent tracker, KCF \cite{KCF}, as the code is available online.
 
After running the one-pass evaluation (OPE), we obtain two types of graphs based on different metrics. The precision plot shows the percentage of frames for which the center location error (CLE) is lower than a Location Error Threshold (LET), with CLE computed as the Euclidian distance between the tracker estimated location and the groundtruth's center. On this plot, trackers are ranked by the precision obtained for LET = 20 pixels.
The second graph is the success plot. It represents the percentage of frames for which the overlap ratio (OR) is larger than a given threshold. This ratio is computed between the intersection and union of the bounding box given by the tracker $(B_T)$ and the groundtruth $(B_G)$ 
\[
OR = \frac{S(B_T\cap B_G)}{S(B_T\cup B_G)}
\]
with $S$ denoting the number of pixels of the covered surface. The ranking on this plot employs the area under the curve (AUC) value as it measures the overall performance instead of the success obtained for a single threshold.

\begin{figure*}[t]
     \centering
     \subfloat[][Overall]{\includegraphics[width = 4.5cm]{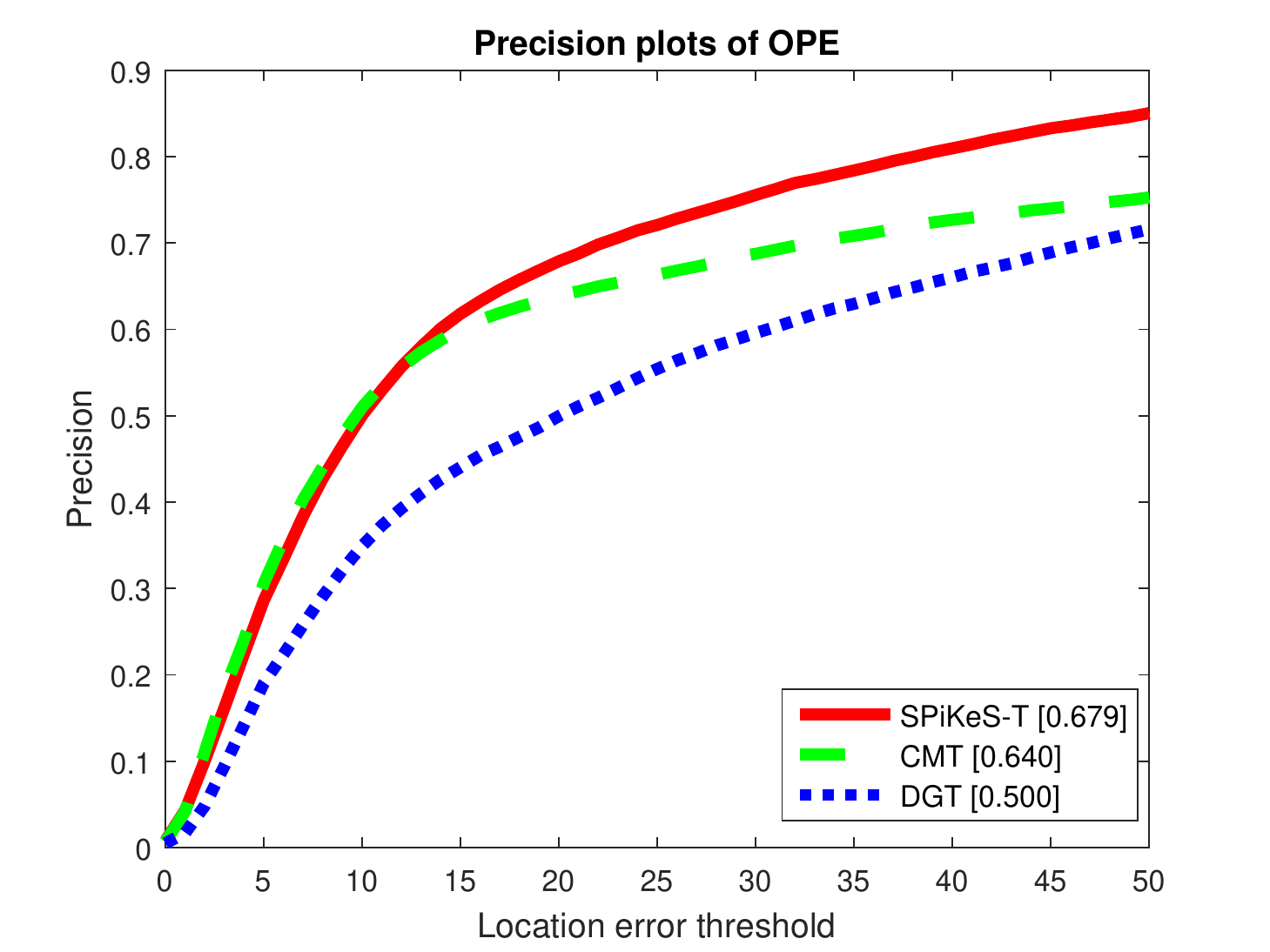} \includegraphics[width = 4.5cm]{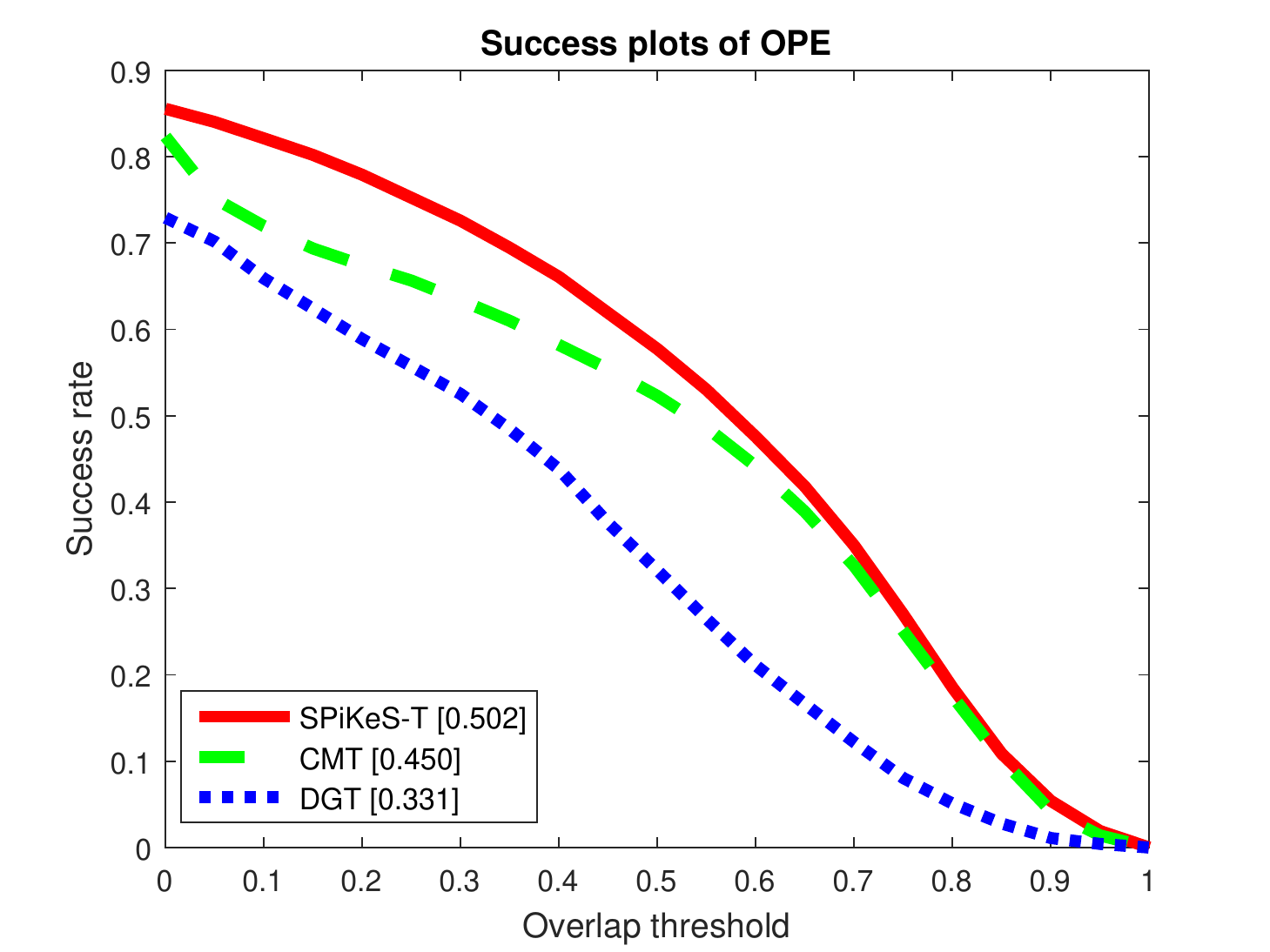}\label{im:Overall2}}
     \subfloat[][Deformation]{\includegraphics[width = 4.5cm]{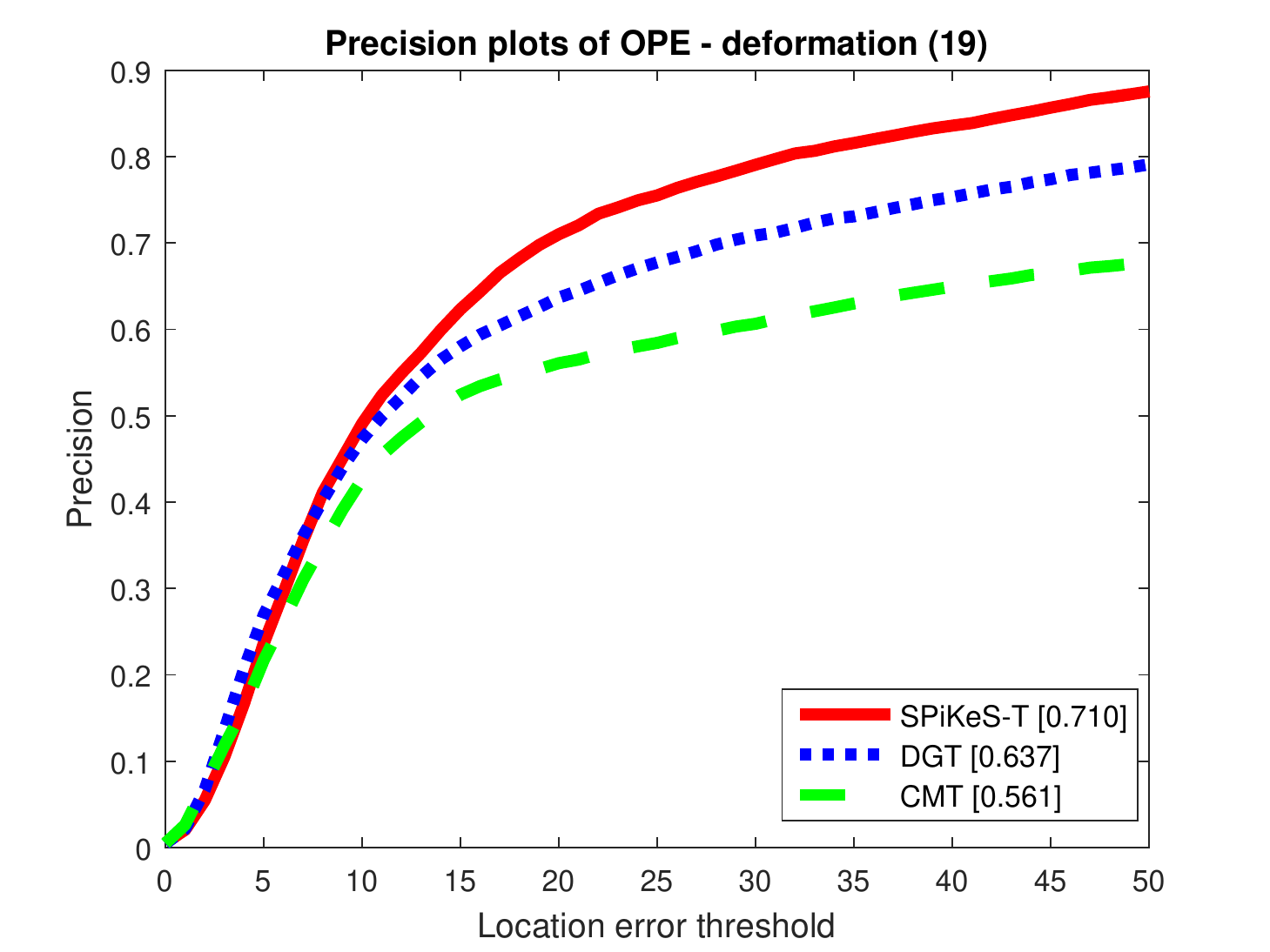} \includegraphics[width = 4.5cm]{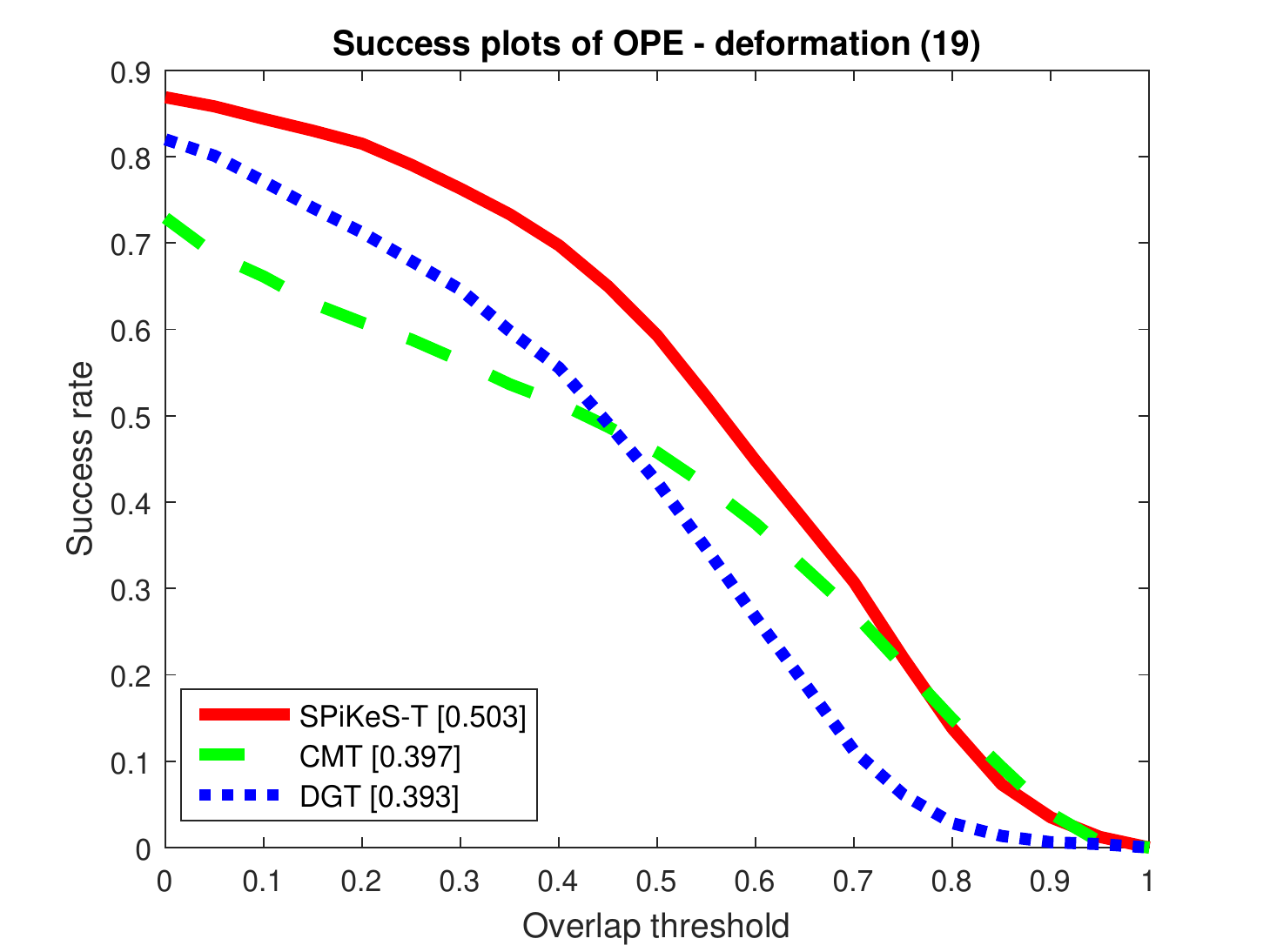}\label{im:def2}}\\
      \subfloat[][Background clutter]{\includegraphics[width = 4.5cm]{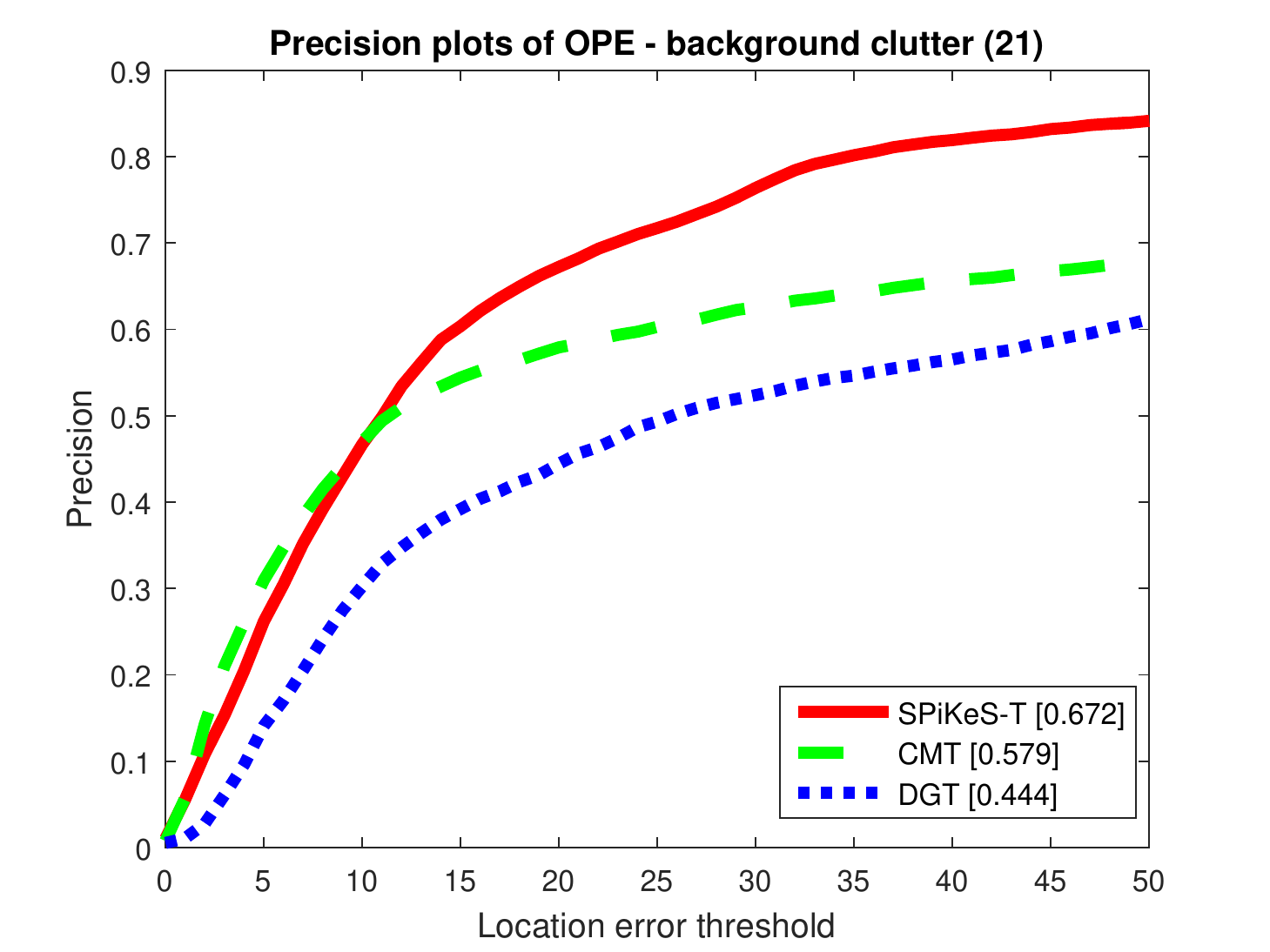} \includegraphics[width = 4.5cm]{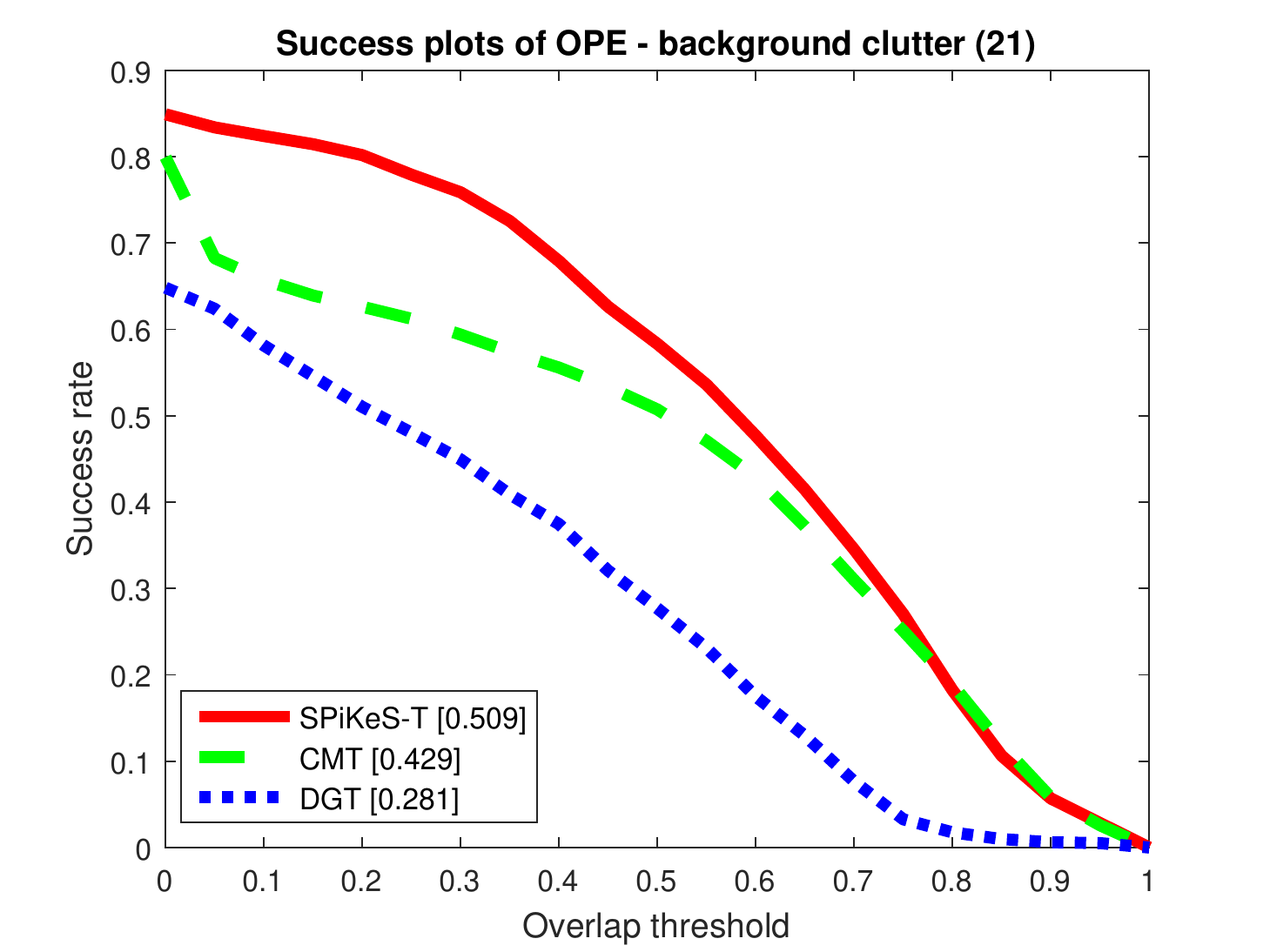}\label{im:bc2}}
     \subfloat[][Occlusion]{\includegraphics[width = 4.5cm]{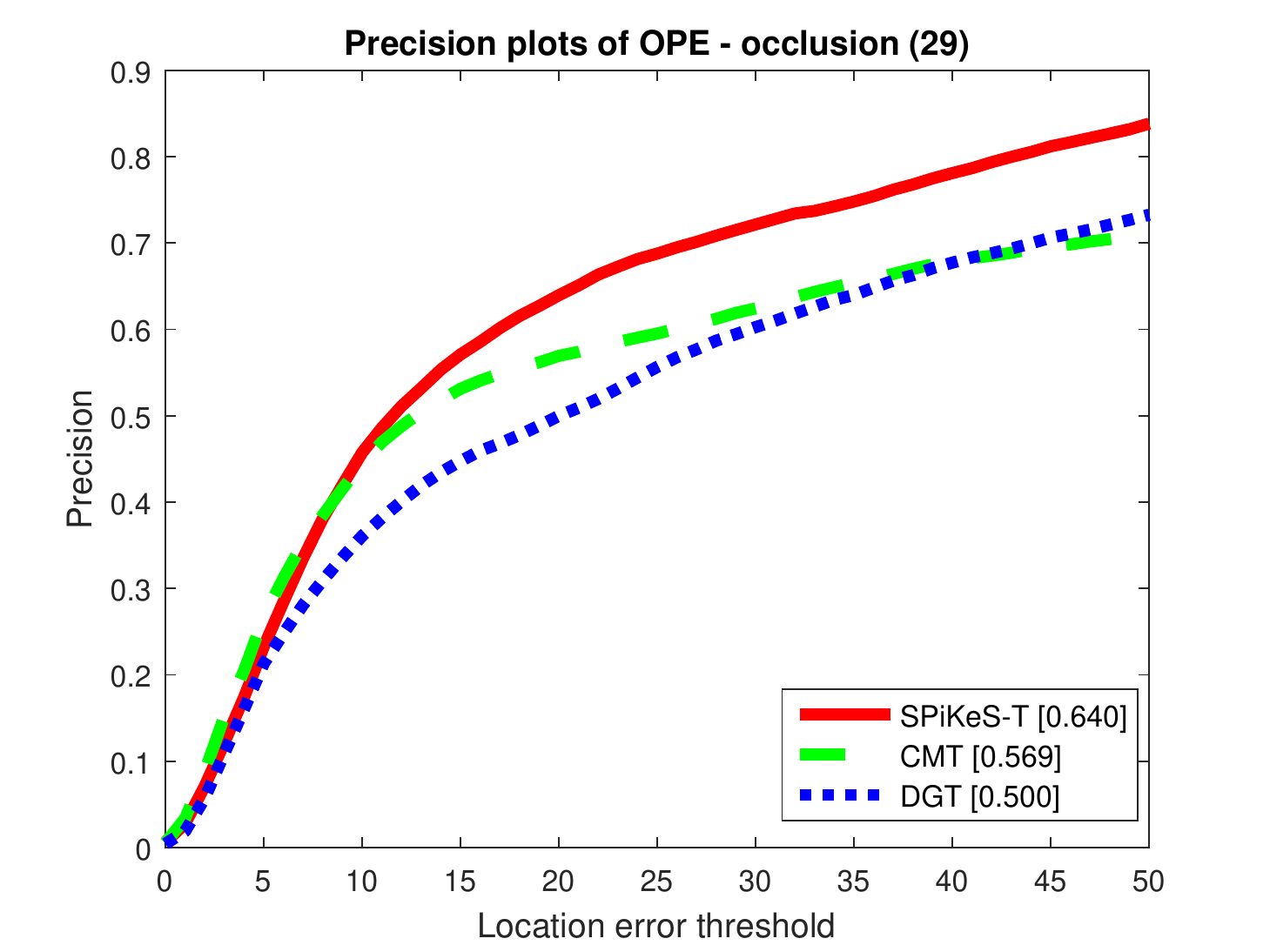} \includegraphics[width = 4.5cm]{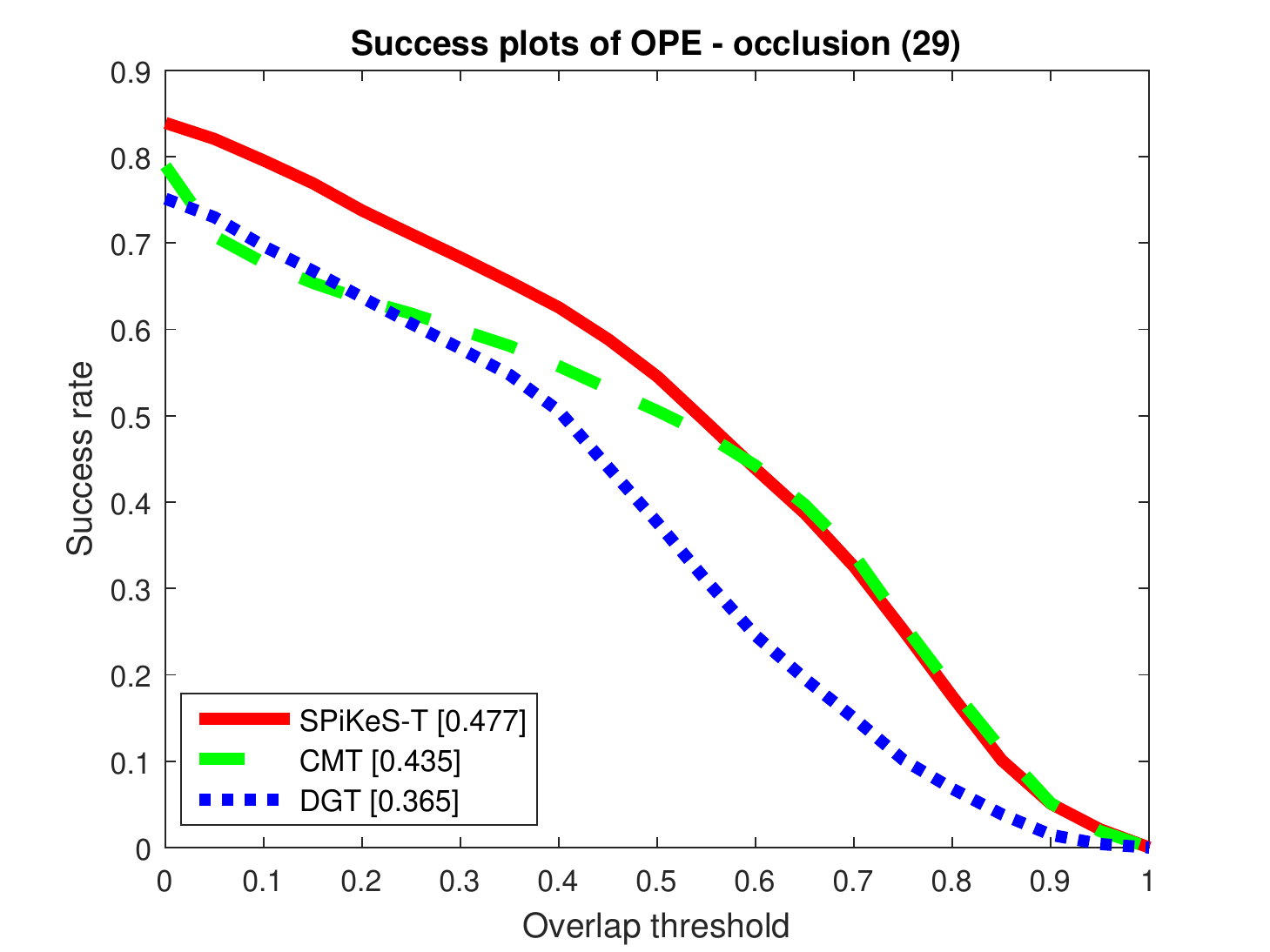}\label{im:occ2}}\\
     \subfloat[][Illumination variation]{\includegraphics[width = 4.5cm]{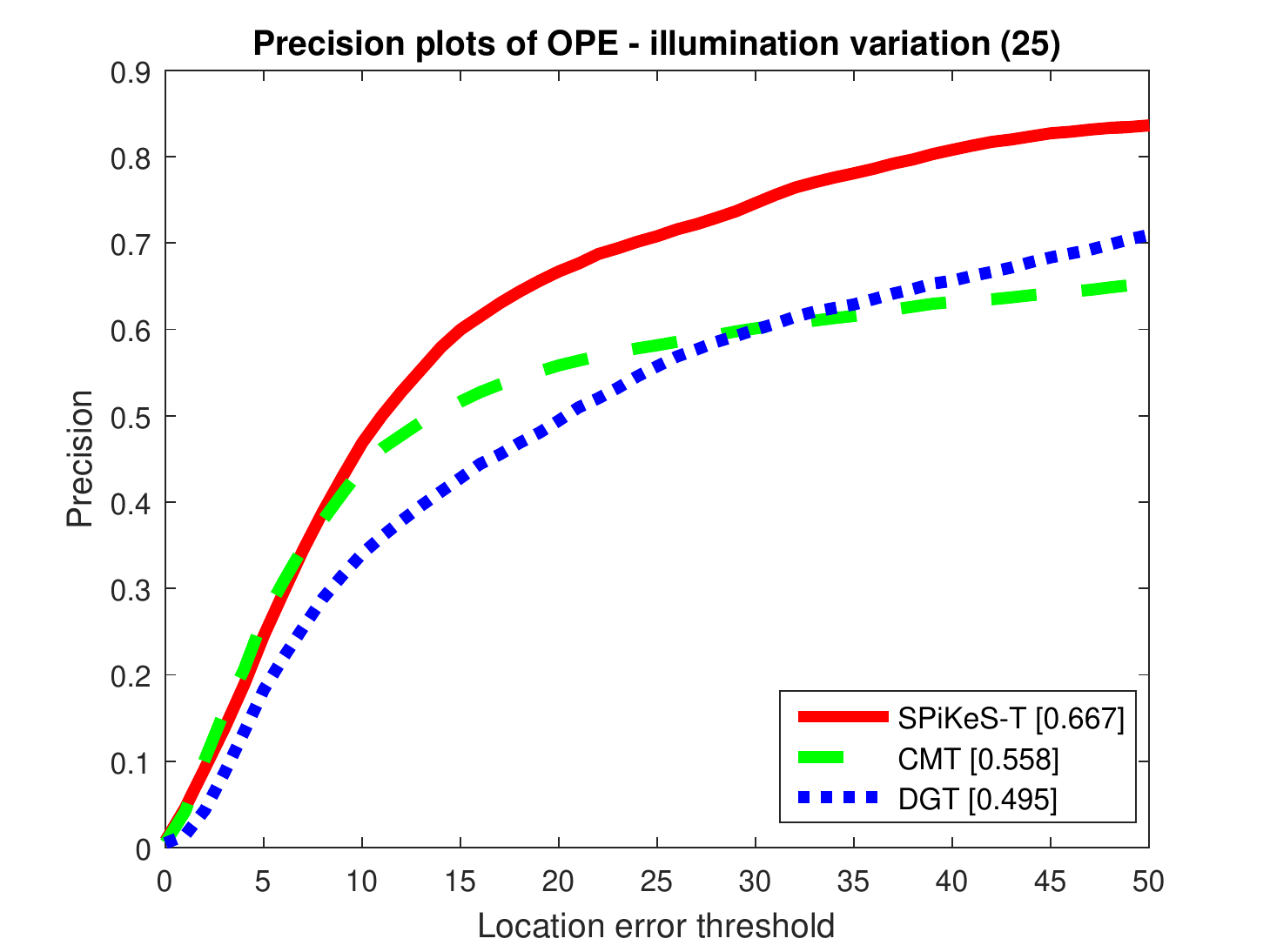} \includegraphics[width = 4.5cm]{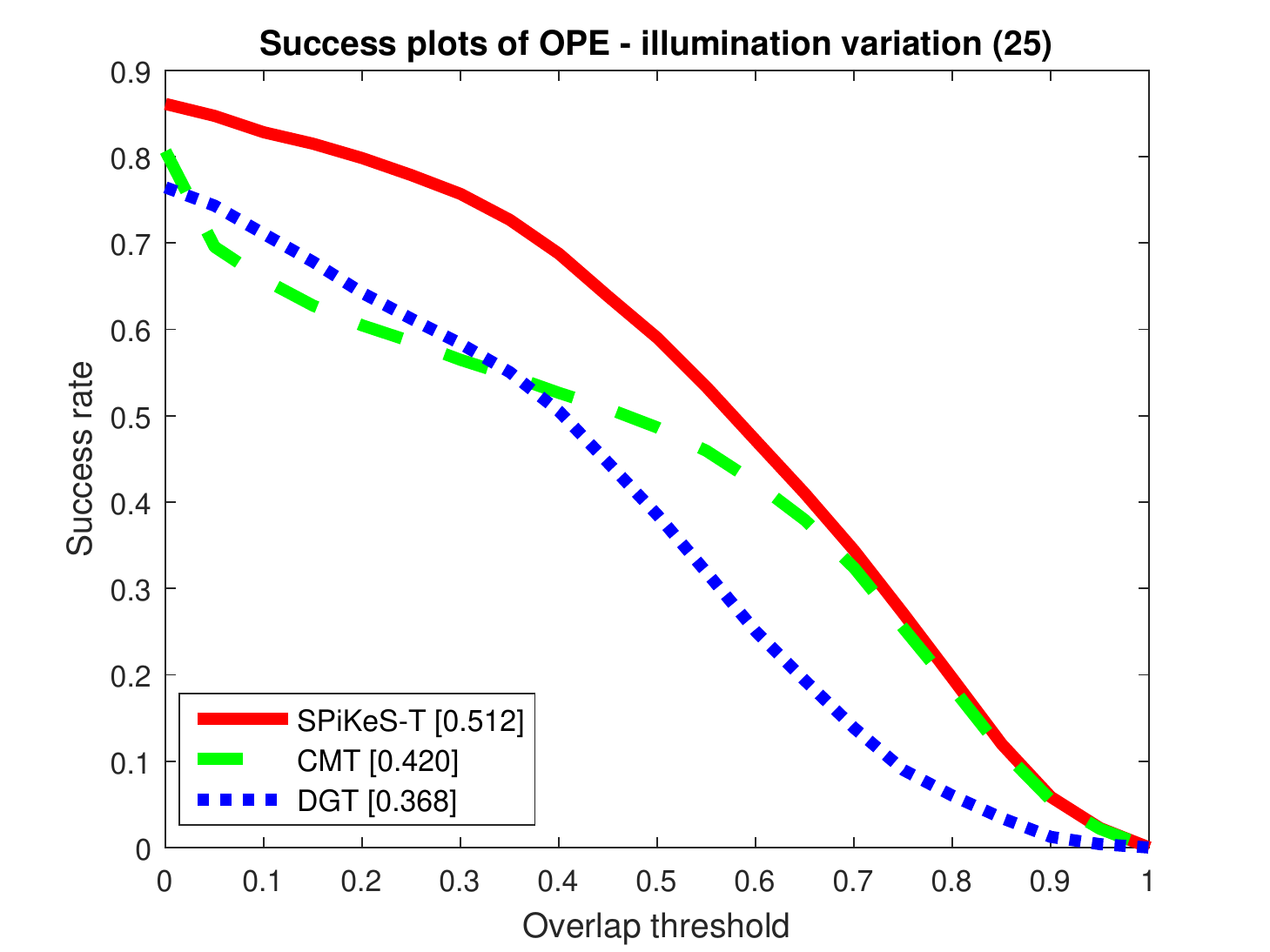}\label{im:iv2}}
     \subfloat[][Fast motion]{\includegraphics[width = 4.5cm]{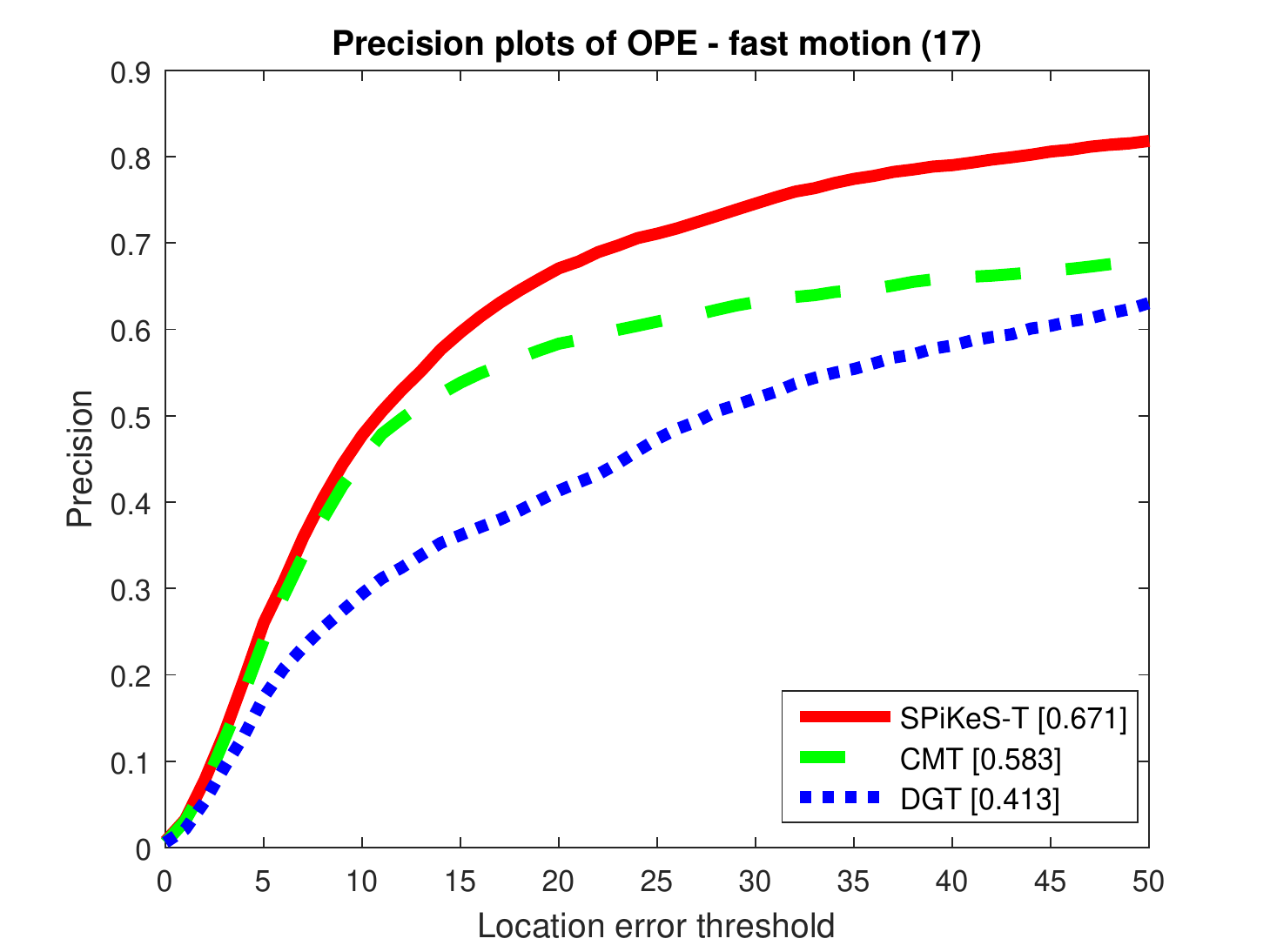} \includegraphics[width = 4.5cm]{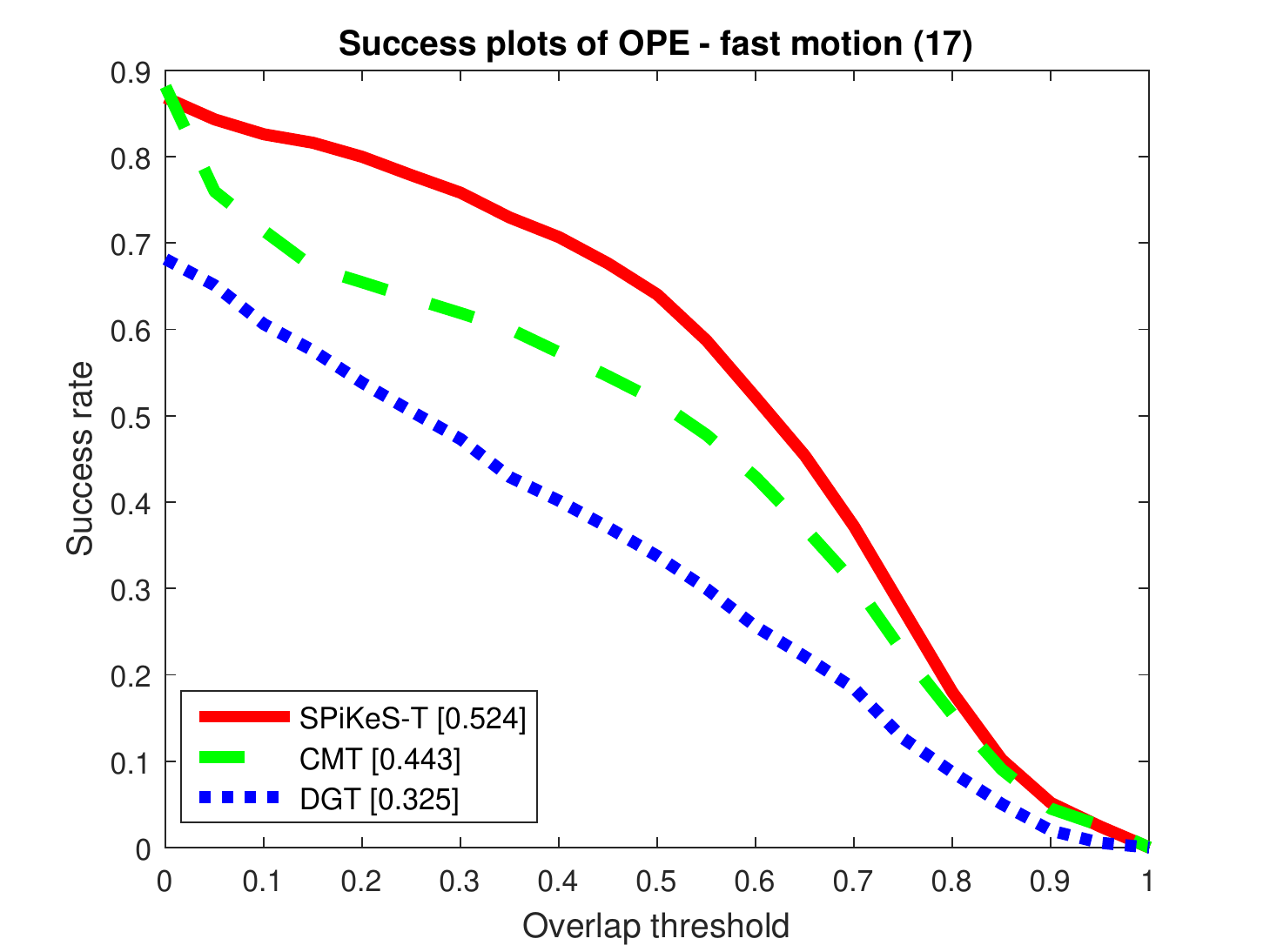}\label{im:fm2}}
     \caption{Comparison of a superpixel-based tracker (DGT), a keypoint-based tracker (CMT) and ours (SPiKeS-T) on OTTB.}
     \label{im:result2}
\end{figure*}

Figure \ref{im:Overall} shows the overall plots obtained from the whole dataset, while plots \ref{im:bc}-\ref{im:occ} are obtained from subgroups gathering videos of the same challenging factor. Only the top ten methods are shown for clarity. We observe that our tracker (SPiKeS-T) gives promising performances since it ranks first for almost all of these cases. However, it reaches only the second place on the overall plot after KCF. KCF has the benefits of a scale adaptation, unlike our method, which explains why it tracks better on sequences with scale variation (fig. \ref{im:sv}).  

\begin{figure*}[]
\captionsetup[subfigure]{labelformat=empty}
     \centering
     \subfloat[][]{\includegraphics[width= 4.4cm,height=2cm]{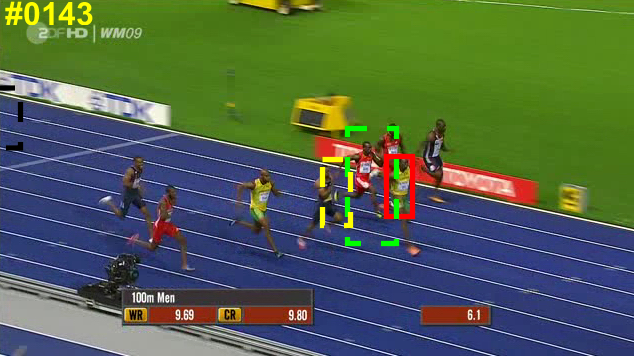}} \thinspace
     \subfloat[][]{\includegraphics[width= 4.4cm,height=2cm]{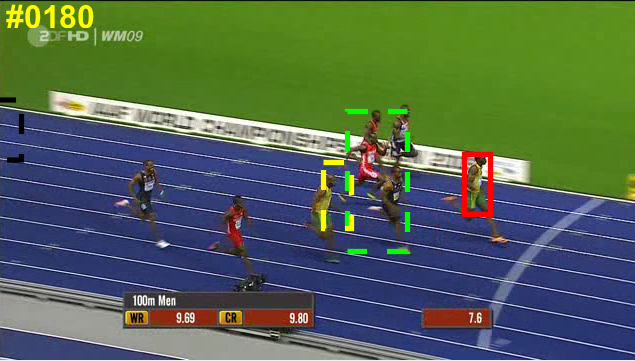}}\thinspace
     \subfloat[][]{\includegraphics[width= 4.4cm,height=2cm]{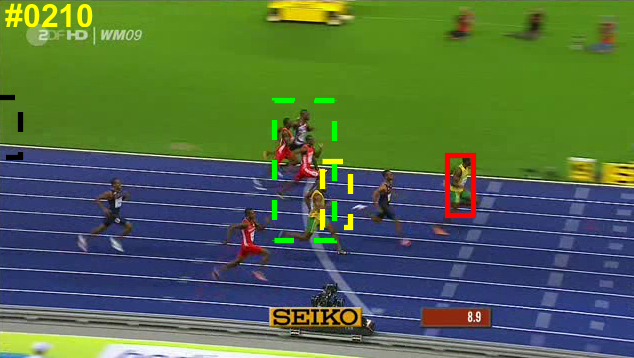}}\thinspace
     \subfloat[][]{\includegraphics[width= 4.4cm,height=2cm]{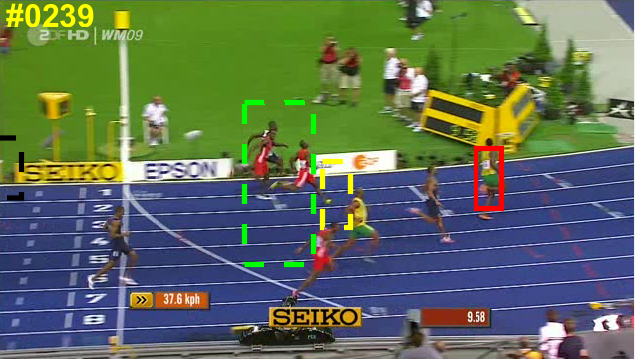}}\\
     \vspace{-1.9\baselineskip}

     \subfloat[][]{\includegraphics[width= 4.4cm,height=2cm]{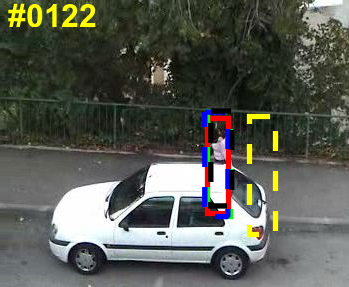}} \thinspace
     \subfloat[][]{\includegraphics[width= 4.4cm,height=2cm]{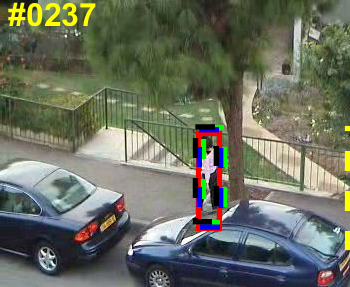}}\thinspace
     \subfloat[][]{\includegraphics[width= 4.4cm,height=2cm]{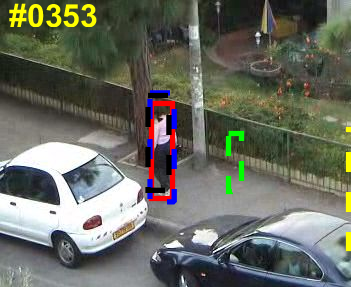}}\thinspace
     \subfloat[][]{\includegraphics[width= 4.4cm,height=2cm]{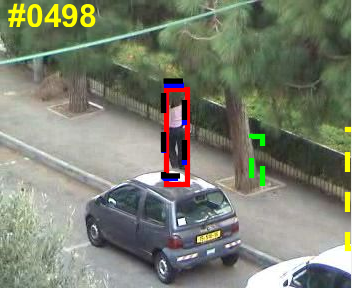}}\\
     \vspace{-1.9\baselineskip}

     \subfloat[][]{\includegraphics[width= 4.4cm,height=2cm]{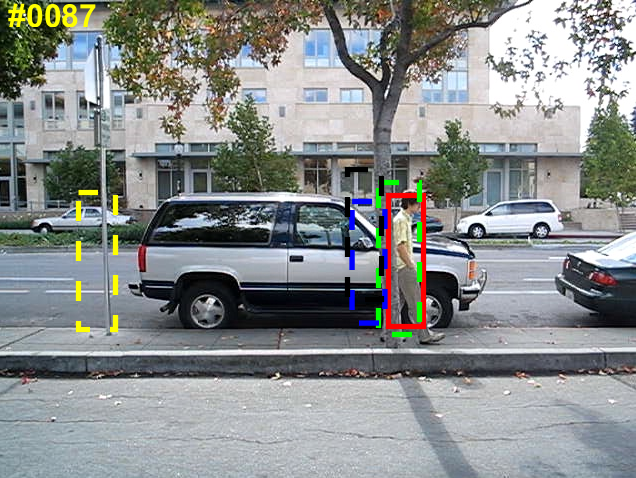}} \thinspace
     \subfloat[][]{\includegraphics[width= 4.4cm,height=2cm]{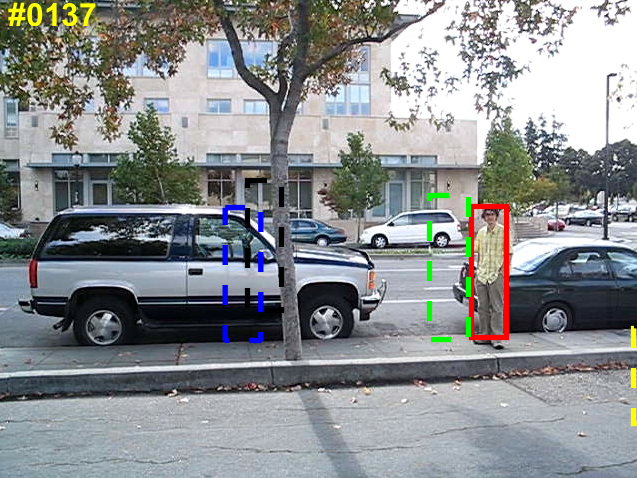}}\thinspace
     \subfloat[][]{\includegraphics[width= 4.4cm,height=2cm]{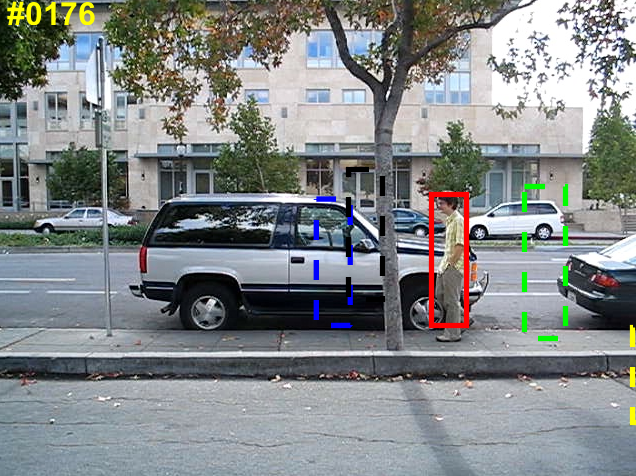}}\thinspace
     \subfloat[][]{\includegraphics[width= 4.4cm,height=2cm]{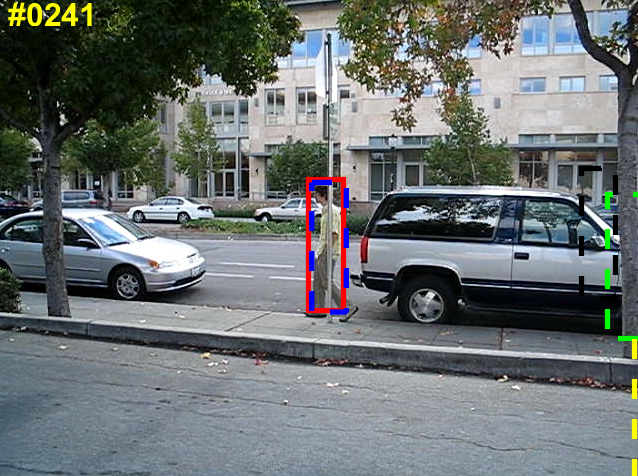}}\\
     \vspace{-1.9\baselineskip}

     \subfloat[][]{\includegraphics[width= 4.4cm,height=2cm]{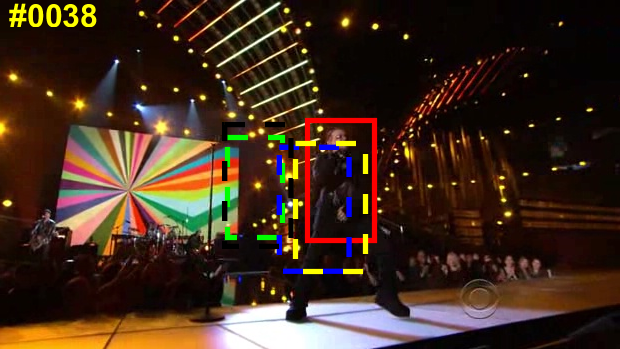}} \thinspace
     \subfloat[][]{\includegraphics[width= 4.4cm,height=2cm]{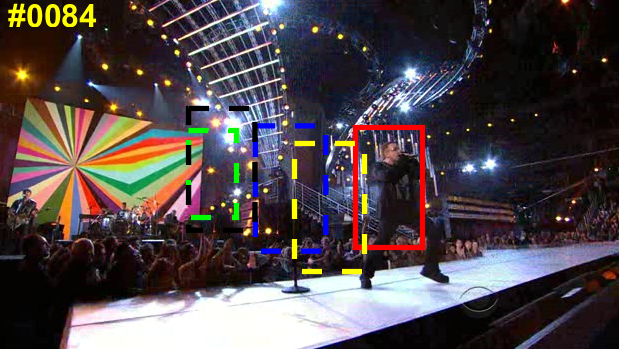}}\thinspace
     \subfloat[][]{\includegraphics[width= 4.4cm,height=2cm]{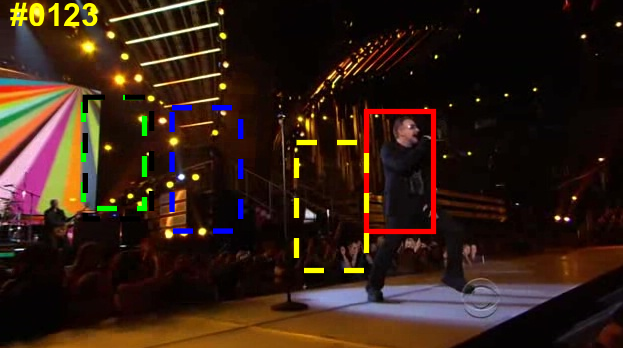}}\thinspace
     \subfloat[][]{\includegraphics[width= 4.4cm,height=2cm]{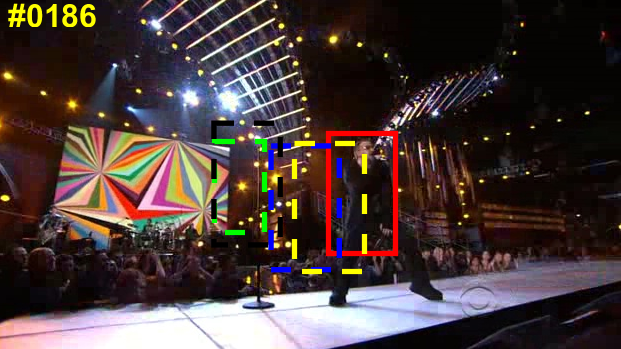}}\\
     \vspace{-1.1\baselineskip}

\subfloat[][]{
    \cbox{red} SPiKeS-T \quad
    \cbox{green} KCF \quad
    \cbox{blue} SCM \quad
    \cbox{black} Struck \quad
    \cbox{yellow} TLD \quad
}	
 \vspace{-1.1\baselineskip}
    \caption{Qualitative results of top five trackers for sequences \textit{bolt}, \textit{woman}, \textit{david3}, \textit{singer2} from top to bottom.}
     \label{fig:QualR}
\end{figure*}

As Struck does not adapt to scale change either, we can fairly compare to it on the overall success and precision plots where our method reaches better performance on both. This is mainly due to our part-based model, which shows best results against deformation as seen in figure \ref{im:def}. Indeed, our local parts, the SPiKeS, are very flexible as we do not enforce rigid connection between them. Each one is matched regardless of the others allowing large deformations. Moreover, a superpixel is a deformable part itself and better represents local deformation. This advantage has been exploited in our update scheme making our tracker more robust against background clutter than the other trackers as we observe on figure \ref{im:bc}. For example, SCM updates its generative model with rectangular patches, which are less reliable than superpixels as patches cannot adapt to the shape. Qualitative results for the top five trackers are also presented on figure \ref{fig:QualR}.

\subsubsection{Comparison to related trackers}
As our goal is to show the benefits of the SPiKeS for tracking, we also compare our method to two specific recent trackers:  CMT \cite{kp_vote2} and DGT \cite{DGT}. Both are part-based trackers which locate their target with votes like ours. However, the former is a keypoint-only tracker while the latter is a superpixel-only tracker. As their codes are available online, we evaluate them on the benchmark of Wu et al. \cite{WuLimYang13}, keeping the default parameters given by the authors. Results in figure \ref{im:result2} show that our approach outperforms the other two, demonstrating that combining superpixels and keypoints leads to a more robust tracking than using these features alone. More specifically, the results can be explained for different situations:

\begin{itemize}
\renewcommand{\labelitemi}{$\bullet$} 
\item \textit{Deformation}: As they are all part-based trackers, they are more suited to handle deformations. To alleviate the lack of discriminativity of superpixels, DGT employs \textit{spectral matching} to match a graph of superpixels. This technique requires the computation of an affinity matrix. However, for that matrix to be computationally manageable, constraints on the deformation must be set. Consequently, this tracker fails in case of heavy deformation. As for CMT, keypoints can be difficult to match when the target undergoes deformation, since some keypoints will disappear and new ones appear. Therefore, only a few matches will determine the location, which will be inaccurate if some of the matches are wrong. On the contrary, as an image can always be segmented in a same number of superpixels, numerous SPiKeS are candidates to be matched even in case of deformation. Moreover, since SpiKeS may have keypoints in common, a single keypoint can lead to several matches of SPiKeS resulting in a more accurate location.

\item \textit{Background clutter}: In this situation, the background distracts the tracking and often leads to a model drift. Where DGT will match wrong superpixels and CMT false keypoints, the more discriminative power of a SPiKeS helps in avoiding such ambiguous matches. Indeed, the color of a superpixel can avoid bad keypoint matches while the structure of keypoints can differentiate two superpixels of similar color. Compared to CMT, the boundary evidence brought by a superpixel avoids adding noisy keypoints to the model, as presented in figure \ref{im:updateSpikes} in the previous section. Furthermore, even if noisy SPiKeS are added to our model, the persistence and prediction factors favor reliable SPiKeS, which also prevents the model from drifting.

\item \textit{Occlusion}: On this curve, we see that DGT is less efficient. If the occluder has similar color as the target, it will be classified as foreground and no occlusion will be detected. Thus, it will not be able to avoid updating the model which will make it drift. To detect occlusion, it seems that keypoints are better but SpikeS has still an advantage over keypoints-only trackers. In case of a missed occlusion detection and an unwanted update, it is less probable to add bad keypoints to a SpiKeS thanks to the boundary of the superpixel. However, as keypoint-only trackers have no clue as whether a new keypoint belongs to the target, it is more likely that the model will drift due to a  background keypoint added erroneously to the model.

\item \textit{Illumination variation}: In case of illumination variation, DGT tends to fail as it relies only on color and unlike CMT that detects BRISK keypoints \cite{brisk}, our tracker uses SIFT keypoints which are designed to be robust against illumination variation.

\item \textit{Fast motion}: Constraints on the affinity matrix  computed by DGT also limits the motion of each of its superpixels. This is why it performs poorly when there is fast motion. Our tracker adapts its motion constraint according to the target's motion.
\end{itemize}

\begin{figure}[!h]
\begin{center}
\includegraphics[width = 4.35cm]{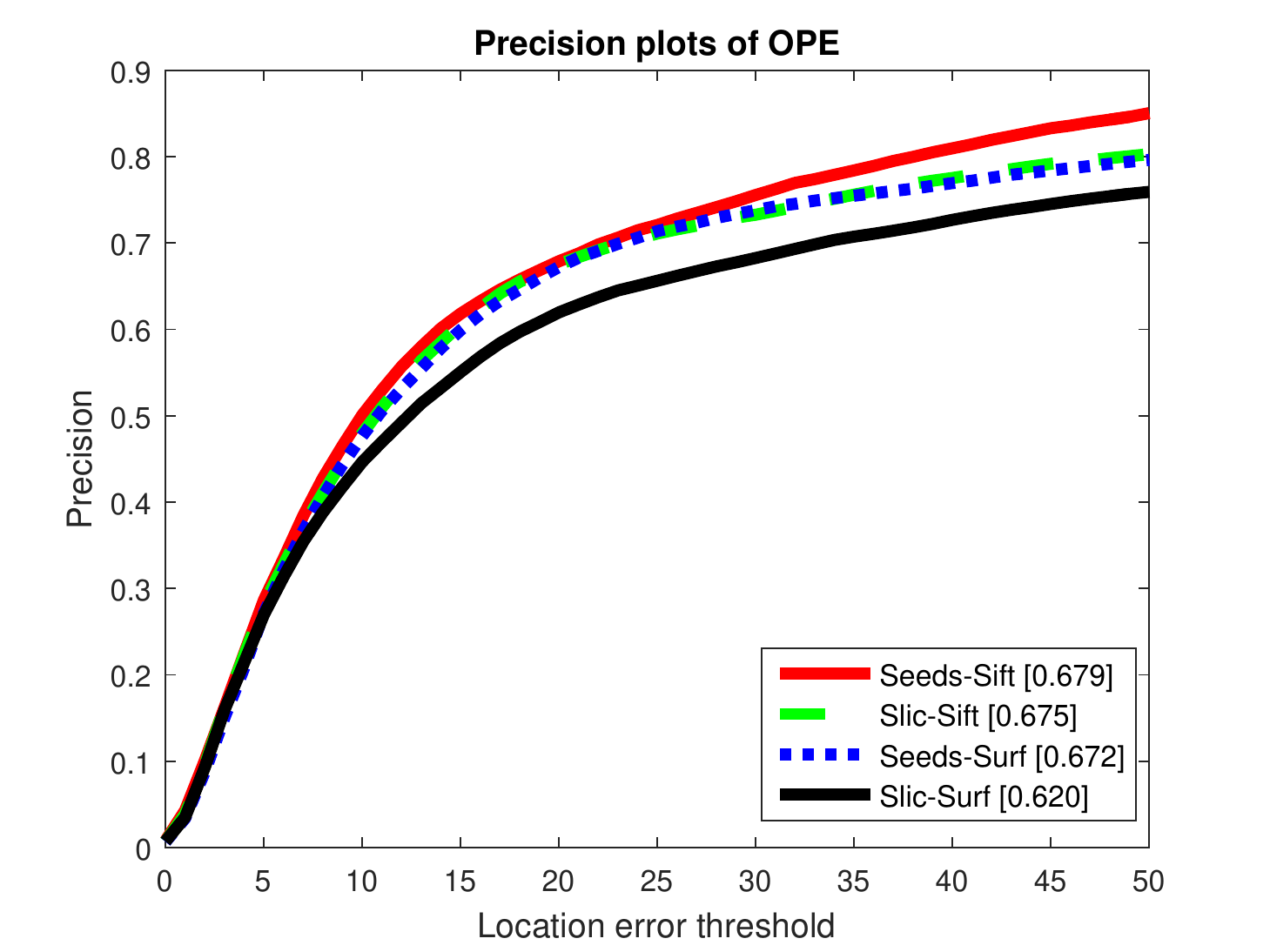} 
\hspace{-0.5cm}
\includegraphics[width = 4.35cm]{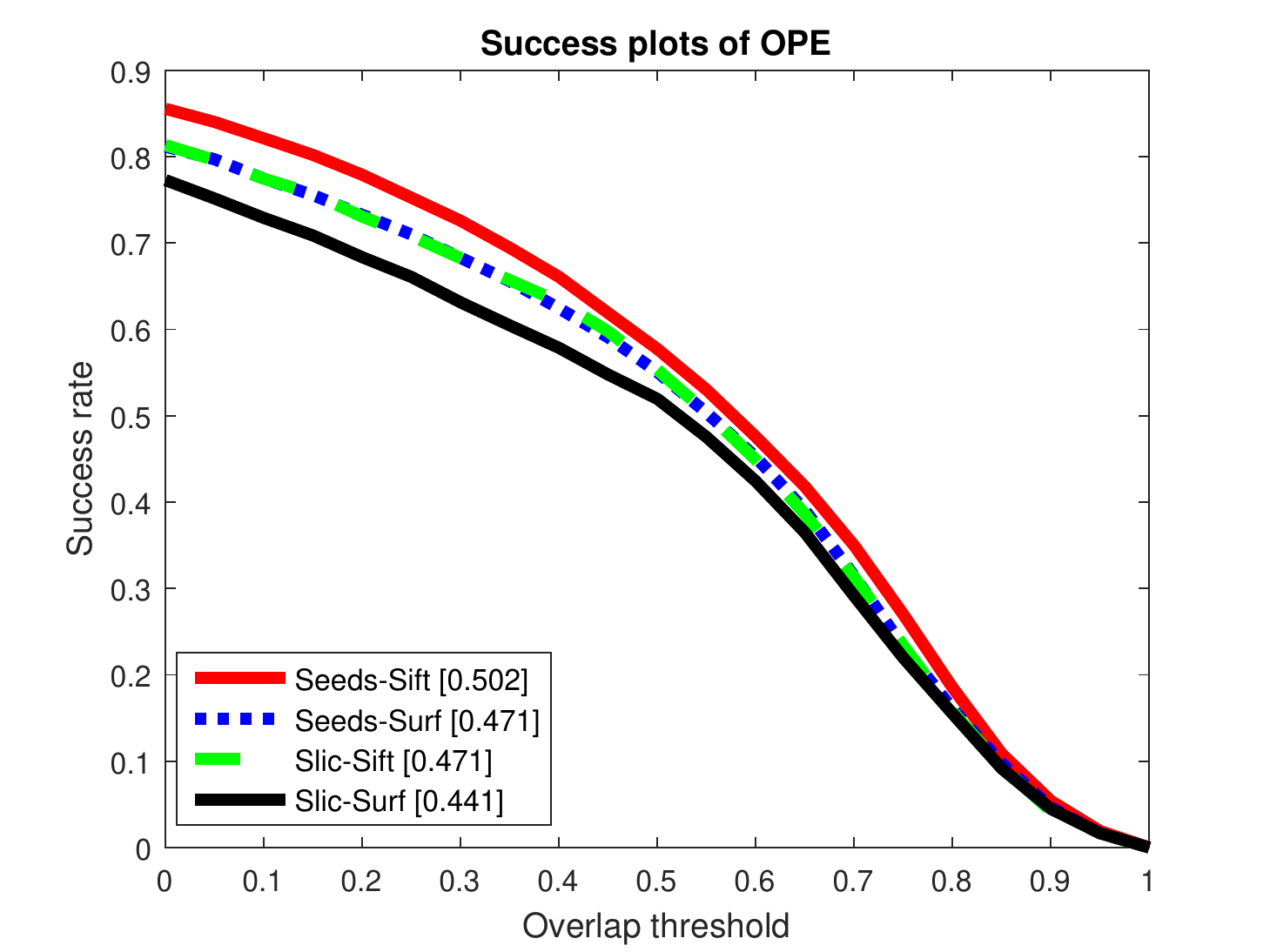}
\end{center}
\caption{Influence of different superpixels-keypoints combinations on the overall performance on OTTB.}
\label{im:featComp}
\end{figure}

It is also interesting to see the influence of other types of superpixels and keypoints to build the SPiKeS. Figure \ref{im:featComp} compares different combinations of features such as SIFT \cite{SIFT} and SURF \cite{surf} for keypoints and SLIC \cite{SLIC} and SEEDS \cite{SEEDS} for superpixels. Although the results are quite equivalent, the best combination is not a surprise. SEEDS has shown to fit better to boundaries than SLIC in \cite{SEEDS} and SIFT is more robust than SURF to illumination changes \cite{sift_surf_comp}.

\section{Conclusion}

In this paper, we proposed a novel feature combining superpixels and keypoints that we called SPiKeS. We showed that this new feature can be matched efficiently by a simple nearest neighbor technique. Therefore, we developed a SPiKeS-based tracker that leverages this matching to locate accurately target parts in a new frame. Furthermore, based on the SPiKeS properties, we provided a reliable update scheme that avoids the model to drift. Finally, the evaluation against the state-of-the-art shows promising results, as our results are close to the ones of KCF tracker, even outperforming it in many scenarios, despite that our tracker does not yet include an adaptation to scale variation. In addition, our superior performance compared to superpixels-only and keypoints-only trackers first demonstrates the benefits of fully combining these two features for more robust tracking. As a final word, we point out that the use of SPiKeS could advantageously be extended to other applications such as object detection and foreground segmentation.

\vspace{1cm}
\par{{\centering
{\bf\uppercase{Acknowledgements}\\[0.5em]}
}}
\noindent This work is supported by Fonds de recherche du Québec - Nature et technologies (FRQ-NT) team grant \#172083. We gratefully acknowledge the support of NVIDIA Corporation with the donation of the Titan X GPU used for this research.

\bibliographystyle{IEEEbib}
\bibliography{SpiKeS_T_paper_bib}
\end{document}